\documentclass{article}
\usepackage[T1]{fontenc}
\usepackage{graphicx}
\usepackage{microtype}
\usepackage{natbib}
\usepackage{pifont}
\let\cite\citep

\usepackage{color}
\usepackage[usenames,dvipsnames]{xcolor}
\usepackage{hyperref}
\hypersetup{
    colorlinks = true,
    allcolors = Cerulean,
    breaklinks = true
}
\usepackage[capitalize]{cleveref}

\definecolor{lightgray}{gray}{0.9}

\usepackage{fullpage}
\usepackage{subcaption}

\author{
Ilia Kuznetsov\textsuperscript{\tiny{1}}\footnote{Corresponding author;
1 -- Technical University of Darmstadt, UKP Lab; 
2 -- Mohamed bin Zayed University of Artificial Intelligence;
3 -- Radboud University Nijmegen;
4 -- Carnegie Mellon University;
5 -- Allen Institute for Artificial Intelligence; 
6 -- Bocconi University, Milan;
7 -- The University of Sydney;
8 -- Universität Hamburg, Data Science Group;
9 -- University of British Columbia;
10 -- Indian Institute of Technology, Delhi;
11 -- University of Applied Sciences Darmstadt;
12 -- Université Paris-Saclay, CNRS, LISN;
13 -- Indian Institute of Science, Bangalore;
14 -- Monash University;
15 -- The Hebrew University of Jerusalem;
16 -- University of Washington;
17 -- Georgia Institute of Technology;
18 -- Queen's University;
19 -- IT University of Copenhagen.
}
\and
Osama Mohammed Afzal\textsuperscript{\tiny{2}}
\and
Koen Dercksen\textsuperscript{\tiny{3}}
\and
Nils Dycke\textsuperscript{\tiny{1}}
\and
Alexander Goldberg\textsuperscript{\tiny{4}}
\and
Tom Hope\textsuperscript{\tiny{5}}
\and
Dirk Hovy\textsuperscript{\tiny{6}}
\and
Jonathan~K.~Kummerfeld\textsuperscript{\tiny{7}}
\and
Anne Lauscher\textsuperscript{\tiny{8}}
\and
Kevin Leyton-Brown\textsuperscript{\tiny{9}}
\and
Sheng Lu\textsuperscript{\tiny{1}}
\and
Mausam\textsuperscript{\tiny{10}}
\and
Margot Mieskes\textsuperscript{\tiny{11}}
\and
Aurélie Névéol\textsuperscript{\tiny{12}}
\and
Danish Pruthi\textsuperscript{\tiny{13}}
\and
Lizhen Qu\textsuperscript{\tiny{14}}
\and
Roy Schwartz\textsuperscript{\tiny{15}}
\and
Noah A.~Smith\textsuperscript{\tiny{16, 5}}
\and
Thamar Solorio\textsuperscript{\tiny{2}}
\and
Jingyan Wang\textsuperscript{\tiny{17}}
\and
Xiaodan Zhu\textsuperscript{\tiny{18}}
\and
Anna Rogers\textsuperscript{\tiny{19}}
\and
Nihar B. Shah\textsuperscript{\tiny{4}}
\and
Iryna Gurevych\textsuperscript{\tiny{1}}
}

\title{What Can Natural Language Processing Do for Peer Review?\footnotetext{This paper is an outcome of Dagstuhl seminar 24052 ``Reviewer No.~2: Old and New Problems in Peer Review'',\\ \url{https://www.dagstuhl.de/24052}.}}

\date{}

\begin{document}

\maketitle

\begin{abstract}
The number of scientific articles produced every year is growing rapidly. Providing quality control over them is crucial for scientists and, ultimately, for the public good.
In modern science, this process is largely delegated to peer review---a distributed procedure in which each submission is evaluated by several independent experts in the field. Peer review is widely used, yet it is hard, time-consuming, and prone to error. Since the artifacts involved in peer review---manuscripts, reviews, discussions---are largely text-based, Natural Language Processing (NLP) has great potential to improve reviewing. As the emergence of large language models (LLMs) has enabled NLP assistance for many new tasks, the discussion on machine-assisted peer review is picking up the pace. Yet, where exactly is help needed, where can NLP help, and where should it stand aside? The goal of our paper is to provide a foundation for the future efforts in NLP for peer-reviewing assistance. We discuss peer review as a general process, exemplified particularly by reviewing at artificial intelligence (AI) conferences. We detail each step of the process from manuscript submission to camera-ready revision, and discuss the associated challenges and opportunities for NLP assistance, illustrated by existing work. We then turn to the big challenges in NLP for peer review as a whole, including data acquisition and licensing, operationalization and experimentation, and ethical issues. To help consolidate community efforts, we create a \emph{companion repository} that aggregates key datasets pertaining to peer review (\url{https://github.com/OAfzal/nlp-for-peer-review}). Finally, we issue a detailed call for action for the scientific community, NLP and AI researchers, policymakers, and funding bodies to help bring the research in NLP for peer review forward. We hope that our work will help set the agenda for research in machine-assisted scientific quality control in the age of AI, within the NLP community and beyond.
\end{abstract}

\section{Introduction}
One can argue that the two of this year's most discussed topics at Natural Language Processing and Artificial Intelligence conferences will be (1) the unprecedented expansiveness of potential applications of our rapidly advancing technology (in spite of many new and remaining open fundamental questions) and (2) the frustration with peer review.
This white paper proposes bringing (1) and (2) together, by arguing that the progress in NLP creates a unique opportunity for the NLP research community to tackle the challenges associated with the scientific review process. How could NLP researchers focus their efforts to improve the effectiveness and efficiency of peer review, and what concrete steps can be taken to this end?  We argue that peer review is inherently interesting to the NLP community as an application area, opening up new problems, revealing new variants of familiar ones, and perhaps inspiring fresh ideas about data, methods, and evaluation.

Peer review is a critical component of sound scientific discovery and, by extension, of the accountability to the public that is increasingly affected by science and its applications. The core peer review process was originally established in research communities of a few hundred people, relatively few of whom were in trainee career stages, and none of whom were subjected to the ``fast science'' pressure to disseminate findings early.  
Given the rapid growth experienced by AI research communities in particular~\cite{kunzli2022not}, and by science as a whole~\cite{landhuis2016scientific}, existing processes do not scale well. 
This leads to increasing frustration with peer review. After months (or years) of hard work, authors expect informed, well-reasoned, unbiased evaluation of their results by qualified people in relatively short period of time.  
Yet reviewing is a complex and time-consuming task, and qualified reviewers are in short supply and overloaded with papers to review. Additional burden is placed on the organizers of the process, who need to deal with finding reviewers, assigning them to the papers, monitoring the process, and resolving conflicts---all of which requires increasing effort as the community and the number of submissions grow.

The scientific community widely recognizes the new challenges faced by peer review. With the emergence of large language models (LLMs), the topic of fully or partially automated peer review keeps re-emerging both within and outside NLP \cite[and others]{liu2023reviewergpt, Biswas2023ChatGPTAT, liang2023can, robertson2023gpt4, d2024marg, drori2024human}. Related work expresses both the hope that AI can help deal with the reviewing crisis in science, and the worries about the potentially grave consequences of automating this crucial stage in scientific work \cite{schintler2023critical}. Our paper contributes to this debate with an informed discussion of the potential contributions of NLP to peer review. As we show, applying NLP to peer review meets a range of serious challenges that need to be addressed \emph{before any approaches to full automation of peer review can even be considered}. 
Yet, peer review is not one isolated action and not a single NLP task. It is a complex process with dozens of tasks, which all require human effort, and are all prone to failure. While some of these tasks might be out of reach, others are closely related to well-known general NLP challenges such as reasoning, scientific content understanding, cross-document modeling and summarization, human and automatic evaluation, mitigating bias, and upholding ethics and privacy. Thus, while full automation might be not feasible, \textit{some individual well-defined problems within peer review can and should be addressed}---with or without LLMs---and addressing them could make peer review more efficient and robust, save valuable researcher time, and increase satisfaction and trust in the peer review process.
We believe that the NLP community plays the key role in addressing this challenge, due to our collective experience with fast community growth and our technical expertise with NLP technologies, their capabilities, and their limitations. Importantly, the social processes underlying peer review can open up new research directions and new connections to other fields of research such as game theory, sociology, psychology, meta-science, and human-computer interaction. Being at the core of the scientific process, the topic is inherently intriguing and extremely challenging, and has potential to draw new talent toward our community.

With this paper, we aim to map out the problem space of NLP for peer review assistance. We discuss peer review as a general process, using reviewing at AI conferences as a running example (\cref{sec:define}). We then dive into each step of the process---from paper submission to publication---and outline core challenges and automation opportunities that emerge along the way (Sections \ref{sec:before}--\ref{sec:after}), based on first-hand experience in reviewing and organizing scientific events, and on the available NLP technology. This is followed by a general discussion of experimental methods, data collection, and legal and ethical aspects of NLP for peer review (Sections \ref{sec:resource_eval_data}--\ref{sec:ethical_legal}), and a practical call for action (\cref{sec:cfa}). The companion repository for this white paper aggregates datasets related to NLP for peer review and welcomes new contributions. We hope that our work will help foster community efforts in developing NLP technologies for machine-assisted scientific quality control that will benefit reviewers, authors, conference organizers, readers---and the scientific process as a whole.

\section{Background} \label{sec:define}
We prepare our discussion by defining peer review and outlining the core challenges of applying peer review for scholarly quality control. This is followed by a detailed walkthrough of peer-reviewing at AI conferences, along which we introduce the necessary terminology used throughout the paper.

\subsection{Defining Peer Review}

Peer review is a general evaluation procedure where the work is assessed by one or more peers who have a comparable expertise to the producers of the work~\cite{lee2013}.
Peer review is commonly used in scenarios that require non-trivial assessment of complex products at scale. Such scenarios span from software development to medicine, from education to grant evaluation, and from employee assessment to scholarly peer review. As our paper is dedicated to the last of these, for brevity, from here on we use ``peer review'' as a shorthand for scholarly peer review. Peer review is the core quality control mechanism in modern science. While not perfect, it is often ``compared with democracy in being the least bad system available'' for evaluating and prioritizing scientific outputs \cite{Smith_2010_Classical_peer_review_empty_gun}. 
The general expectations of peer review are largely shared. 
At its best, peer review should gate-keep against methodologically flawed research, help prioritize the rest based on novelty and impact, and provide useful feedback to authors \cite{10.12688/f1000research.11369.2,shah2022surveyextended}. Furthermore, reviewing should be objective, thorough, and impartial~\cite{lee2013}. At its realistic minimum, peer review should keep out papers with obvious flaws and help moderating discussion of ongoing work in the community \cite{rogers-augenstein-2020-improve}. 

\subsection{Common Challenges} 
As scholarly peer review faces the realities of modern research, challenges emerge:

\begin{description}
    \item[] \textbf{Scale and logistics:} Science accelerates, and more submissions create a need for more reviews. For example, at present, major AI conferences such as AAAI regularly attract over \textit{ten thousand} submissions each, and each of these submissions needs to be reviewed, discussed, and decided upon. 
    Coordinating these massive efforts poses new logistical challenges for the organizers of the peer-reviewing process, such as maintaining the technical infrastructure capable of dealing with the high workload, assigning reviewers to submissions, moderating the reviewing process, and making fair acceptance decisions.
    
    \item[] \textbf{Cost:} Reviewing takes time: reading the draft, doing background literature research, writing a helpful review report, engaging in discussion with the authors and other reviewers, \emph{and} potentially repeating the process in the revise-and-resubmit cycle all draw on the limited time resources of the community. Estimates  repeatedly arrive at \emph{millions of human-hours} spent reviewing yearly \cite{aje, gspr}---time not spent collecting data, running experiments, writing papers, studying, or teaching. Given that peer review is performed by experts, the economic cost of these human-hours is also considerable.
    
    \item[] \textbf{Bias:} Peer review is a hard, subjective task. The growth and diversification of research communities introduce new potential sources of bias in peer review: prestige~\cite{peters1982peer,murray2016bias,tomkins2017,manzoor2021uncovering}, nationality~\cite{ernst1991chauvinism, 10.1001/jama.295.14.1675}, gender and race~\cite{goldberg1968women,ginther2011race,regner2019committees, 
    strauss2023racism}, and language proficiency~\cite{10.1001/jama.295.14.1675} can all influence peer review outcomes, along with confirmation bias (valuing submissions that conform to the reviewer's beliefs), publication bias (valuing positive over negative results), and others~\cite[Sections 6 and 7]{shah2022surveyextended}. While there exist various bias mitigation strategies like reviewer and author anonymization or assigning multiple reviewers per submission, these leave the problem far from solved.
    
    \item[] \textbf{Low-quality reviewing:} Increasing submission rates make it necessary to lower the barrier of entry for reviewing, with more and more junior researchers joining the reviewer pool---especially in rapidly growing fields like machine learning and NLP. Given the increasing specialization of research topics, even an experienced reviewer will not have equally deep expertise in every research area. The combination of lacking qualification, hard task, and time pressure can lead to low-quality reviewing. 
    This includes both generic, hastily written reviews, and the use of fast-reject heuristics like ``not state of the art'', ``too niche'', or ``writing too bad'' instead of a thorough evaluation (e.g., \url{https://aclrollingreview.org/reviewertutorial}). While there exist organizational measures to improve reviewing quality, such as reviewer mentoring~\cite{stelmakh2020novice} and training courses\footnote{Such as  \url{https://clarivate.com/web-of-science-academy} or \url{https://github.com/reviewingNLP/ACL2020T3material}}, low-quality reviewing remains a persistent issue. 
    
    \item[] \textbf{Strategic and dishonest behavior:} Finally, peer review relies upon reviewer impartiality. Yet, by definition, reviewers are themselves researchers, and they often have their own work under review for the same venue. Given the competitive nature of modern research, reviewers have an incentive to abuse their role. This includes downscoring others' work to reduce potential competition, establishing collusion rings in which a group of reviewers conspire to pick each others' work for review and give it a favourable evaluation (Section \ref{sec:colrings}), and other behaviors~\cite[Section 4]{shah2022surveyextended}.
\end{description}

Together, these factors can make peer review less successful at achieving what it is designed to do. A lot is at stake: passing peer review gives manuscripts the special status of a 
``peer-reviewed scientific publication'', which often---for the better or worse---plays a large role in formal evaluations (\textit{``$x$ peer-reviewed papers to get a PhD''}), 
researchers' standing within the field (\textit{``over 100 papers in A* venues''}), and perception of the work both by fellow researchers and by non-experts and media (\textit{``a peer-reviewed article has shown that...''}). Mistakenly approved bad work can misguide follow-up research and misinform policymaking and public opinion. Mistakenly rejected good work, in turn, delays the dissemination of findings, causes unnecessary resubmission effort, and has a direct impact on researchers' careers. This motivates the ongoing search for policies, incentives, and tools that can help peer review stay robust to the realities of modern research. Our paper contributes to this important line of work by outlining the ways in which state-of-the-art NLP technology can support peer review.

\subsection{Walkthrough: Reviewing at AI Conferences}

Science is diverse, and so is peer review. While distant research communities might have similar expectations and face similar challenges related to peer review, their specific practices and workflows can widely differ. To structure the discussion throughout this paper, we focus on a particular use case most familiar to the primary audience of this paper: conference peer review in AI-related computer science communities. 
Example communities include NLP (e.g., ACL, NAACL, EMNLP), computer vision (e.g., CVPR, ICCV, ECCV), general AI (e.g., AAAI, IJCAI, UAI) and general machine learning (e.g., NeurIPS, ICML, ICLR). Although there are differences between specific conferences (and even between different editions of the same conference), the underlying peer-reviewing process is by and large similar. We now describe this process in detail, while introducing necessary terminology along the way. We note that our description is \textit{modular}, and many challenges and solutions outlined in this paper will be applicable to a wider range of reviewing systems, e.g., journal review and open post-publication review. We also note that while particular artifacts and roles might be specific to AI conferences, most concepts involved are \textit{discipline-agnostic} and can be easily adapted to other research communities and non-academic use cases.

\begin{figure}[t]
\centering
\begin{subfigure}[t]{0.5\textwidth}
        \centering
        \includegraphics[height=2.5in]{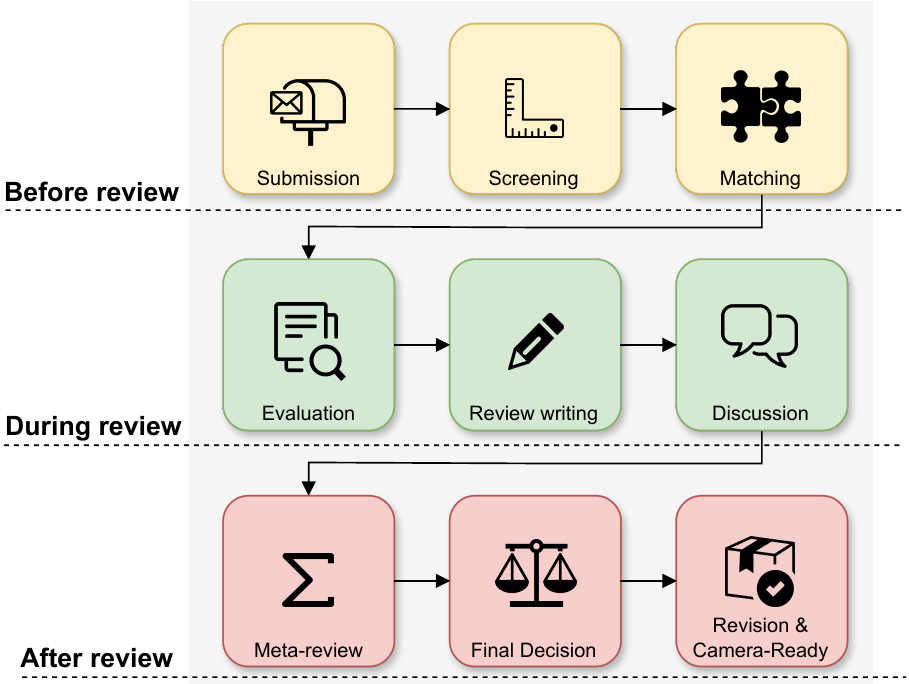}
        \caption{Key stages of peer review}
        \label{fig:pr_a}
    \end{subfigure}%
    ~ 
    \begin{subfigure}[t]{0.4\textwidth}
        \centering
        \includegraphics[height=2.5in]{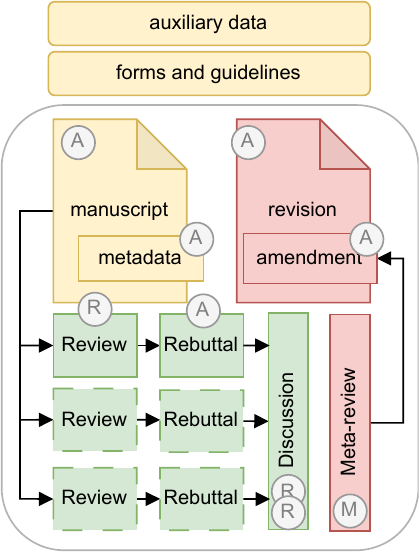}
        \caption{Artifacts}
        \label{fig:pr_b}
    \end{subfigure}
    \caption{Peer review as a process, along with the resulting artifacts. Stages of peer review color-coded.\\(A) -- authors, (R) -- reviewers, (M) -- meta-reviewers.}
\end{figure}

Following the initial call for papers, the organizers of an AI conference initiate a \emph{reviewing campaign}, illustrated in Figure \ref{fig:pr_a}. Reviewing is managed by a program committee which is headed by \textit{program chairs} (PCs). They assemble the reviewer pool by recruiting reviewers---other members of the community who often submit manuscripts themselves. PCs are responsible for setting up the technical infrastructure for peer review, using \emph{conference management systems} (CMS) like SoftConf (\url{https://softconf.com}), Microsoft CMT (\url{https://cmt3.research.microsoft.com}), and, increasingly, OpenReview (\url{https://openreview.net}). In addition, PCs decide on the review forms and guidelines to be used in the campaign, including the scoring system, checklists, and submission rules. Program chairs are supported by \textit{meta-reviewers} (sometimes also called area chairs), and---at larger conferences---by \textit{senior meta-reviewers} (also called senior area chairs). Some conferences additionally enlist a dedicated ethics committee. Prior to the start of the reviewing process, the participants---authors, reviewers, and meta-reviewers---are often requested to fill out profiles that contain information about their affiliation, topics of expertise, prior publications, conflicts of interest, etc.

In the meantime, \emph{authors} prepare their \emph{manuscripts} (papers), which are submitted to the conference via the CMS at a due date along with metadata such as keywords, track and contribution type. AI conferences typically follow a double anonymized peer review model: by design, the authors and the reviewers do not know each others' identities.\footnote{At some conferences the author identities are known to the PCs and meta-reviewers. Pre-printing and advertising the work on social media are not uncommon either
, potentially making the authors non-anonymous~\cite{rastogi2022arxiv}.} 
Upon submission, the manuscript goes through semi-automatic screening to ensure that it adheres to basic criteria like formatting, length limit, and anonymity. Failing this step might result in desk rejection without review. If a submission passes the formal checks, it is assigned to a set of \emph{reviewers} (typically 3--5) from the reviewer pool. The assignment is done via semi-automatic matching, bidding, or a combination thereof. This step aims to maximize topical overlap and diversity of the reviewer set for each manuscript, while avoiding conflicts of interest. 

Then, reviewing begins. The reviewers independently evaluate their assigned manuscripts and each of them writes a \emph{review report}: a short semi-structured essay evaluating the work, often accompanied by a range of scores (e.g., overall score, confidence, soundness, novelty) and checklists (e.g., adherence to ethics guidelines or data availability). The reviewing workload varies, typically 3--6 submissions per reviewer. The evaluation is often followed by discussion, which consists of two parts. In the author-response phase (also known as \emph{rebuttal}) the authors see their papers' reviews and can respond to reviewers' questions and concerns. This is followed by the reviewers \emph{discussing} the manuscript among themselves to arrive at the final version of their review reports. Once the reviewing is done, the review reports, author responses and discussions serve as input to the meta-reviewers who can additionally provide their own judgement of the work. Meta-reviewers may request further reviews in case of missing or low-quality reports, or if there are ethical concerns about the manuscript. Following that, meta-reviewers write a \emph{meta-review report}---a brief text that summarizes reviewers' feedback, and provides their own evaluation of the work---typically 10--20 manuscripts per meta-reviewer. 

At large conferences, based on the manuscript, its reviews and its meta-reviews, as well as any considerations of target acceptance rates, senior meta-reviewers then give a \emph{recommendation} on whether or not to accept the paper to be published in the conference proceedings---around 20--50 manuscripts per senior meta-reviewer. Finally, all papers, meta-reviews and recommendations are passed to the program chairs---2--4 very senior members of the community responsible for the conference program. They make the final acceptance/rejection decisions for all manuscripts submitted to the conference. 
If a manuscript is accepted, the authors prepare a \emph{camera-ready revision} of their submission that includes the necessary revisions and is later published. If the manuscript is rejected, the authors often resubmit their work to another venue, ideally incorporating reviewers' feedback into the new version of the manuscript. The revision is oftentimes accompanied by \emph{amendment notes} that summarize the changes. Once the process is complete, conference organizers can analyze the rich data resulting from the reviewing campaign  
to inform future review organization and to provide insights to the community.

\section{Scope of this Paper}

As we have discussed, peer review serves as a crucial quality control mechanism in modern science. It is a complex process that involves many independent actors. As science accelerates, peer review faces new challenges. Which of these challenges are solvable, and to what extent can AI help? We note that \emph{texts} are central to the peer-reviewing process: submitted manuscripts largely consist of text, and so do the reviewing guidelines, organizers' communications, peer review reports, rebuttals, discussion comments, meta-reviews, and the resulting publications along with their amendment notes (Figure \ref{fig:pr_b}). To a large extent, \textit{reviewing work is text work}, and some parts of it bear resemblance to annotation work \cite{rogers-augenstein-2020-improve}. Thus, NLP techniques could be leveraged to support peer review. As we shall see, many of the challenges in NLP for peer review are special cases of general NLP challenges familiar from other domains. Peer review provides both a new testing ground for approaches that emerge elsewhere, and can serve as a unique source of new tasks, data, and methods that can benefit other application areas. 
The goal of this paper is to help consolidate the efforts in this space and to provide an entry point for researchers and practitioners interested in using natural language processing to help peer review. 

The rest of the paper is structured as follows. We start with a detailed discussion of NLP support for peer review, organized around three temporal stages of a peer-reviewing campaign (Figure \ref{fig:pr_a}). \mbox{\textbf{Before review}}, NLP can provide assistance with preparing the submission (\cref{sec:before_prep}), reviewer--paper scoring (\cref{sec:before_matching}) and matching (\cref{subsubsec:matching}). \textbf{During review}, NLP can assist in evaluation (\cref{sec:during_eval}), review writing (\cref{sec:during_writing}) and during discussion (\cref{sec:during_discussion}). \textbf{After review}, NLP can help perform meta-reviewing (\cref{sec:after_meta}), assist decision making by the PCs (\cref{sec:after_decision}), support manuscript revision (\cref{sec:after_revision}) and facilitate post-review analysis (\cref{sec:after_post}). While we do \textit{not} systematically survey all prior research in this space, we outline core application points and challenges in NLP for peer review and illustrate the discussion by existing works. We then turn to overarching challenges in NLP for peer review, including collecting and securing data (\cref{sec:resource_eval_data}), measurement and experimental methodology (\cref{sec:resource_eval_eval}), and ethical issues that accompany peer-reviewing applications of NLP (\cref{sec:ethical_legal}). We conclude with an explicit call for action for the authors, reviewers, NLP and AI researchers, policymakers, and funding organizations (\cref{sec:cfa}).

We note that while this work proposes many possible applications of NLP for peer review, each of them hinges on \emph{open research questions}; while we do not argue that the proposed solutions are yet practical or even possible, we do believe that they are worthy of exploration. We also note that while our text is organized by stages of the review process, many of the discussed solutions afford \emph{multiple uses}: for example, a reviewer can use screening tools (\cref{sec:before_screening}) at evaluation stage (\cref{sec:during_eval}) to check a manuscript for misconduct; a meta-reviewer (\cref{sec:after_meta}) can use review and discussion analysis tools (\cref{sec:during_discussion}) to detect low-quality reviews; and any analytical tool can be used for aggregate statistics after a reviewing campaign is over (\cref{sec:after_post}). Thus, even a single NLP assistance approach at a particular stage can benefit a large number of people and the process as a whole. Finally, while we discuss ethics- and data-related questions since they directly influence the NLP practice, our paper does not focus on policies, incentives, or other organizational measures to improve peer review, which constitute an important topic of their own. 

\section{Assistance Before Review}
\label{sec:before}

\begin{figure}[t]
\centering
\includegraphics[width=\textwidth]{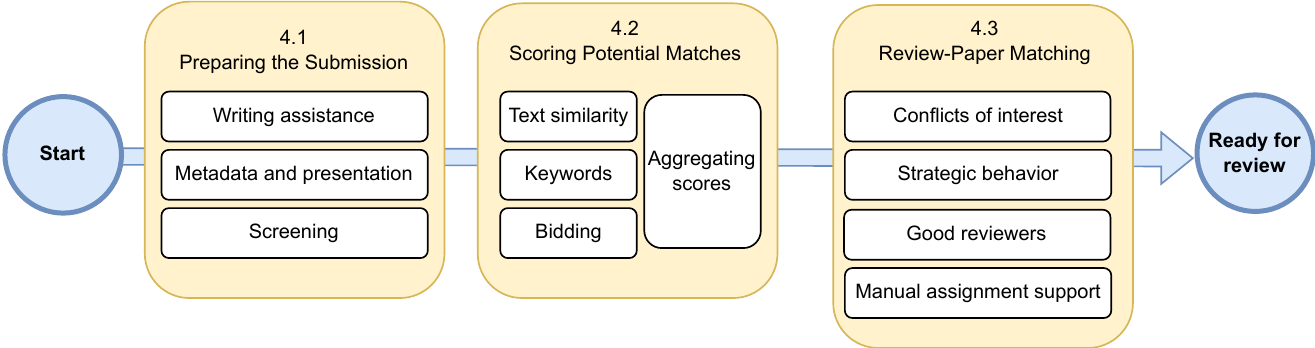}
\caption{Areas of assistance before peer review.}
\label{fig:before_review}
\end{figure}

The preparation phase is crucial for a successful reviewing campaign, and we start our overview by discussing the key steps that precede the actual reviewing: preparing the manuscript for submission, identifying potential reviewer candidates for the submission from the reviewer pool, and making the reviewing assignments. We note that this list is not exhaustive and much more needs to be done before reviewing begins---like recruiting the initial pool of reviewers or issuing a call for papers. Here we focus on the tasks internal to the reviewing process that we believe to be the most promising targets for NLP assistance (Figure \ref{fig:before_review}).

\subsection{Preparing the Submission}
\label{sec:before_prep}

Efficient and robust assessment of scientific work starts with a clearly written and appropriately filed manuscript. Clarity issues can make manuscripts harder to read and evaluate, steering the precious resource of peer review away from the main scientific content of the submission. Even with stellar scientific content, a poorly composed paper is more likely to be rejected and later resubmitted \cite{Church_2020_Emerging_trends_Reviewing_reviewers_again}, incurring additional work for both authors and future reviewers. 
The goal of NLP assistance at this stage is to help make submissions easy to review, while reducing preparation effort for the authors. In addition, NLP can provide assistance during revision and resubmission of the manuscripts: this case is separately addressed in \cref{sec:after_revision}. 

\subsubsection{Writing assistance} 

Tools to support academic writing are abundant, although currently there exist no ready-to-use solutions that can do it all \cite{AltmaeSola-LeyvaEtAl_2023_Artificial_intelligence_in_scientific_writing_friend_or_foe}. While an in-depth discussion of academic writing assistance is orthogonal to the topic of peer review and lies beyond our scope, we outline the general directions of writing support compatible with the
current norms on use of generative AI in paper writing (e.g., \citealp{rogers2023acl}). Generally, technologies that introduce new intellectual content to the paper are not allowed, since they bear risk of plagiarizing ideas or parts of text from published sources. However, using NLP systems to improve writing and clarity without altering the intellectual contribution of the manuscript is allowed. This includes performing basic surface-level analysis, fixing typos, correcting grammar, flagging issues with verbosity, readability, or repetitions.  
Such functionality is already offered by many commercial tools such as Grammarly (\url{https://www.grammarly.com}) and WordTune (\url{https://www.wordtune.com}), as well as specialized academic writing support systems like Curie (\url{https://www.aje.com/curie}), PaperPal (\url{https://paperpal.com}), and Writefull (\url{https://www.writefull.com}). Many such systems also offer a paraphrasing functionality, which currently focuses on applying scientific writing style to the text rather than improving the structure and organization of a draft. 

In the future, NLP systems could be used to detect common structural issues within technical papers, such as using a symbol or abbreviation before it is defined, or symbol overloading. They could suggest additional forward or backward references (e.g., to figures, definitions, and sections) that aid clarity. Further along the complexity spectrum, and closer to the area of scholarly document processing~\cite{sdp-2020-scholarly,sdp-2021-scholarly,sdp-2022-scholarly}, NLP tools could recommend related papers that should be cited
in a given manuscript \cite{AliKefalasEtAl_2020_Deep_learning_in_citation_recommendation_models_survey, FarberJatowt_2020_Citation_recommendation_approaches_and_datasetsa, narimatsu-etal-2021-task}.
Of course, these recommendations will need to be carefully vetted by the authors. An even more complex task is scientific claim verification \cite{wadden-etal-2020-fact,wadden-etal-2022-scifact}: support for checking the statements in the manuscript against provided or cited evidence. After the NLP tool identifies the unsupported statements, the authors would be suggested to either strengthen the evidence or weaken the claim. Such tools would help the authors and arguably result in stronger submissions. 

\subsubsection{Metadata and presentation}

Along with the manuscript, the authors are often required to provide submission metadata such as keywords, track, and contribution type. This metadata is important: keywords are often used in reviewer--paper matching (\cref{sec:before_matching}) and later in conference program navigation; the track sets the context both for the review and for the later presentation of the accepted manuscripts; and the contribution type helps reviewers with evaluation. Yet, deciding on some of this metadata can be challenging. For example, the ACL-2023 conference featured 26 distinct tracks, and AAAI-2024 used 14 primary and 282 secondary keywords as metadata. Often, a submission would not directly fit any of these tracks and keywords. While some venues allow the authors to add arbitrary keywords to their submissions, these need to be concise yet descriptive, and distilling a paper to a good set of keywords requires thought and knowledge of the field. NLP-assisted keyword generation \cite{caragea-etal-2014-citation, cano-bojar-2020-two} and automatic track suggestion could make the submission process and the subsequent review more efficient.

Additionally, the submission stage could feature assistance for reformatting the submission for alternative presentation modes. Examples include TL;DRs (informal, one-sentence summaries, see \citealp{syed-etal-2018-task,cachola-etal-2020-tldr,mao-etal-2022-citesum}), graphical abstracts~\citep[e.g.,][]{10.3389/frma.2021.719004}, and video previews\footnote{For example, \url{https://uist.acm.org/2024/cfp/\#papers}}. Such materials are used to make the final conference program easier to navigate, are shown in the user interfaces for reviewers and chairs, and are published as part of a centralized effort to promote conference submissions via conference social media accounts. Assistance with adjusting a given work to other presentation modes can additionally speed up the submission process for the authors. Research problems of interest include generating first-draft posters \cite{posterbot}, presentations \cite{ppsgen,doc2ppt}, and accompanying videos given a manuscript. Furthermore, AI could be used to generate an overarching figure for the paper (e.g., based on methods like \citealt{belouadi2023automatikz}), which would help reviewers and future readers to quickly get a gist of the paper. Such generative applications may significantly aid the authors in writing strong papers and subsequently presenting their work.

\subsubsection{Screening}
\label{sec:before_screening}

Before a submission enters the review process, it undergoes a range of additional checks to make sure it adheres to the standards and requirements of the venue. Failing to pass these checks can lead to desk rejection without review. Authors could use screening tools pre-submission to ensure that the manuscript complies with formal requirements, and to minimize the overhead for the reviewers and program chairs.

At a bare minimum, screening should ensure correct manuscript formatting. This includes basic parameters such as page count, font size, layout and margins, correct display of figures, and formatting of references. Tools like ACL pubcheck\footnote{\url{https://github.com/acl-org/aclpubcheck}} offer some of this functionality and can be extended. 
Multimodal tools for surface-level analysis of manuscripts such as detecting image manipulation (e.g., \url{https://www.proofig.com}) can be further deployed, but fall outside of our NLP-centered scope. 

Next, a submission can be evaluated with respect to the content-level editorial requirements. Several AI conferences have recently introduced compulsory discussions of limitations and broader ethical impact of the submissions;\footnote{\url{https://aclrollingreview.org/cfp}} simple NLP automation would be sufficient to detect the presence of these sections in the submission. 
In a similar vein, some checklists\footnote{\url{https://aclrollingreview.org/responsibleNLPresearch}} ask authors to indicate the sections of their papers in which certain issues are discussed, e.g., reproducibility efforts, compensation to human participants, adherence to data licenses \cite{DodgeGururanganEtAl_2019_Show_Your_Work_Improved_Reporting_of_Experimental_Results,NeurIPS_2021_Paper_Checklist,rogers2021just}. Automated checks for whether the checklist answers correspond to the indicated paper sections could help authors avoid accidental errors, speed up checklist completion and facilitate reviewing~\cite{liu2023reviewergpt}. 

Finally, the screening stage should detect violations of general reviewing and publishing rules and policies. This includes checking submissions for plagiarism and self-plagiarism, detecting anonymity violations (\textit{``as shown in our prior work''}), dual submission (to different venues) and duplicate submission~(to the same venue), detecting papers from fake paper mills~\citep{else2021fight}, as well as detecting potentially machine-generated text and enforcing compliance with the editorial guidelines on the use of AI for writing assistance. Initial manuscript quality plays a large role in the subsequent process, motivating the deployment of advanced assessment tools already at the screening stage. For example, NLP assistance could help analyze the submissions for coherence, citation coverage, and readability (see \cref{sec:during_eval}). Such analysis would help program chairs and meta-reviewers identify submissions that could be manually reviewed as candidates for desk rejection and mentoring programs, as well as help balance the reviewing load and study the effects of manuscript quality on the subsequent process even if desk rejection does not take place.  

\subsection{Scoring Potential Reviewer--Paper Matches}
\label{sec:before_matching}

Once a manuscript is submitted and has passed the screening stage, it needs to be assigned to reviewers. Reviewer--paper matching is a crucial step: it is at this stage that both the qualification gap and strategic behavior are best mitigated.  In this section, we discuss a key precursor to matching that offers considerable scope for improvement via NLP techniques: determining how well and in what ways a given reviewer is qualified to review a given submission. Then, \cref{subsubsec:matching} discusses considerations that arise in the identification of optimal matches given additional constraints.

\subsubsection{Text-similarity scoring} NLP-based approaches to reviewer-paper scoring typically focus on deriving a similarity score between a text describing a submission (title, abstract, or full paper) and the publication history of a candidate reviewer. Publication history is commonly extracted from third-party aggregators like Semantic Scholar (\url{https://www.semanticscholar.org}), Google Scholar (\url{https://scholar.google.com}), or DBLP (\url{https://dblp.org}). 
One well-known example of a system employing a text-similarity-based approach is the Toronto Paper Matching System \cite{CharlinZemel_2013_Toronto_Paper_Matching_System_automated_paper-reviewer_assignment_system}, which originally used similarity based on word counts and a latent Dirichlet allocation topic model. Recent ACL conferences have employed another system based on an encoder trained on Semantic Scholar abstracts \cite{wieting-etal-2019-simple,2021_ACL_Reviewer_Matching_Code}. An alternative approach is to derive similarity scores via general-purpose paper representations \cite{cohan2020specter}. 
However, evaluations of current similarity-scoring algorithms have found significant room for improvement~\cite{stelmakh2023gold}. Indeed, while intuitive, similarity-based scoring faces several challenges. First, the publication history of a potential reviewer does not necessarily reflect their expertise or their current interests---and at a large conference it is virtually guaranteed that some reviewers will have incomplete and inaccurate profiles. 
Second, a text similarity score might capture criteria that are irrelevant for reviewer assignment, e.g., choice of notation, discussion of basic related work, phrasing in social impact sections, or stylistic elements. Finally, most modern semantic similarity measures are based on dense representations, and thus lack interpretability~\cite{kim2023assisting}. These factors lead even the NLP community itself to accord low trust to similarity scores as an assignment criterion \cite{thorn-jakobsen-rogers-2022-factors}. Addressing these issues within the text similarity paradigm leaves much room for improvement in future work.

\subsubsection{Keywords} A second way of scoring potential reviewer--paper matches draws on manually provided keywords. For example, ACL-2023 experimented with simple matching by keywords selected by authors and reviewers from a pre-defined list, to permit matching according to area, contribution type, and language studied in the paper. Subsequent analysis~\cite{rogers-etal-2023-report} showed that this approach led to higher than average acceptance rates for papers with ``minority'' contribution types, which would likely otherwise be assigned to people not interested in this type of contribution. 
 Keywords are also more interpretable than text-based similarity scores, and can serve as explanations for the matches. 
Yet, a poor selection of keywords will fail to adequately differentiate large groups of submissions and reviewers, which in turn will impact the quality of the subsequent matching. Assisting conference organizers in preparing effective keyword lists is a promising target for NLP assistance---for example, one could use lists of session titles assigned in the program at the previous iterations of a conference as a starting point. Keyword suggestion faces additional challenges due to the need to disambiguate unrelated keywords and connect the related keywords: for example, the authors might chose ``human factors~:~ethics'' as metadata for their submission, while the reviewer may choose ``philosophical foundations~:~human subject experiments''. This challenge may be mitigated by imputing fractional keyword scores to each non-chosen keyword, e.g., based on co-occurrence data between keywords chosen by other papers and authors~\cite{leytonbrown2022matching}. Further improvements to such techniques can come from using more advanced NLP approaches that directly incorporate keywords' underlying semantics and allow free-form text in place of pre-selected keywords.

\subsubsection{Bidding}\label{sec:before_matching_bidding} At some AI conferences, reviewers can actively ``bid'' by indicating their interest and expertise (or lack thereof) in reviewing specific submissions. 
Bidding should be understood as a measure of reviewer interest rather than expertise; for example, bidders might be more likely to bid on submissions that they believe will be high quality, justifying the effort of reading them carefully. They also may bid more actively in areas where they would like to \textit{develop} expertise, rather than in areas where they \textit{already have} the expertise but no longer actively work. 
Furthermore, as the number of submissions increases, bidding faces limitations: there are many papers to bid on, and the reviewers need to enter many bids. 
These problems can be mitigated if bidding is combined with keyword-based and text-similarity-based approaches, as reviewers can be offered the opportunity to bid only for papers for which they would otherwise be judged to have expertise. This approach could further be augmented to leverage information about the number of bids each paper has already received, aiming to ensure that each paper receives an adequate number of bids~\cite{fiez2020super, meir2020market}.

A further, significant problem with bidding is that it increases opportunities for strategic behavior. Reviewers can bid in order to favorably or unfavorably review work to which they are methodologically aligned or opposed. Reviewers can further use bids to favorably or unfavorably review work they suspect was authored by individuals they count as friends or competitors. Finally, reviewers can engage in explicit \emph{collusion rings}, wherein each member of the ring bids for papers from the other ring members regardless of expertise, ultimately submitting positive reviews regardless of submission quality. Collusion rings have become a topic of grave concern because of recent reports of their increasing prevalence at major computer science conferences \cite{Vijaykumar2020Architecture, littman2021collusion}. Collusive bidding can be mitigated by constraining the ways that reviewers can bid; e.g., each reviewer could be required to bid positively on at least 20 papers, under the assumption that the minimum required bid size is larger than the number of the colluding papers. Collusion can also be curtailed by constraining the matches; see \cref{subsubsec:matching}. As for detecting collusions, algorithms based on bidding data alone fail to perform well~\cite{jecmen2022tradeoffs,jecmen2024detection}. Exploring the use of NLP techniques to analyze review texts and associated metadata, alongside bid analysis, presents a valuable research direction for detecting collusion rings.

\subsubsection{Aggregating scores}

Text similarity assessment, keyword matching, and bidding can complement each other. However, this raises the additional challenge of how they should be combined. AI conferences take different approaches in this regard.  
ACL-2023 mainly relied on keywords, falling back on similarity scores when keywords were unavailable, and then on manual adjustments by the PCs where necessary \cite{rogers-etal-2023-report}. NeurIPS-2016 employed a fixed formula~\cite{shah2018design}. AAAI-2021 aggregated scores from two similarity-based systems and keywords augmented with imputed values, using a manually configured function to derive an aggregate score that takes the bidding choices into account \cite{leytonbrown2022matching}. 
 In general, the choice of the best method for combining the individual scores is not clear, and PCs commonly use heuristics to decide on the combination method. A more principled approach to choosing the assignment method is to use post-hoc evaluations of counterfactuals~\cite{saveski2023counterfactual}. Specifically, one can evaluate the potential quality of counterfactual methods---methods which were not used---based on the data from the reviews obtained. A key question herein is the measurement variable. While current post-hoc evaluations rely on reviewers' self-reported expertise or meta-reviewer ratings of reviews, a much more nuanced assessment could be enabled by merging NLP-based evaluation of reviews with causal inference techniques~\cite{saveski2023counterfactual}.

\subsection{Reviewer--Paper Matching}
\label{subsubsec:matching}

Given scores for every reviewer--paper pair, the next step is to determine the optimal matching that respects the load preferences of the reviewers. This problem can be addressed with discrete optimization techniques \cite{CharlinZemel_2013_Toronto_Paper_Matching_System_automated_paper-reviewer_assignment_system, kobren19localfairness, payan2021will, stelmakh2018forall, rogers-etal-2023-report, leytonbrown2022matching}, given a set of constraints and optionally an objective function to score the matches. 
It is a critical step, and it can materially improve matching quality and consequently the quality of the overall peer review process. The role of NLP techniques is to assist with collecting the information that serves as input for discrete optimization. Such information includes scores, as well as constraints like conflicts of interest, reviewer history, and venue- and community-specific constraints that might need to be derived from the submission data.\footnote{For example, the target language of analysis is an important variable in NLP. At ACL-2023 most submissions focused on English, and relatively few focused on other languages. ACL-2023 aimed to ensure that such submissions had the first choice of reviewers speaking those languages, so as to provide such work with more fair and higher-quality reviews. This had a positive effect---subsequent analysis showed that reviews from reviewers matched by (non-English) language were 1.29 times less likely to be flagged by the authors for review issues~\cite[p.~lxiii]{rogers-etal-2023-report}.}

\subsubsection{Identifying conflicts of interest}\label{sec:coi} One major class of matching constraints are the conflicts of interest (COI), which prohibit a given reviewer from being assigned to a given paper. Most conferences ask reviewers to specify their COI manually, but this approach can fail if reviewers are forgetful, have too many conflicts, or strategically choose to under-report their COIs. It is thus desirable to infer conflicts automatically. Conflicts arising from co-authorship can be inferred if reviewers are required to specify their DBLP profile, Semantic Scholar profile, or ORCID (\url{https://orcid.org}). 
AAAI-2021 applied a system of this kind and found at least one unreported conflict for 78.8\% of submissions---over 96,000 unreported conflicts in total \cite[Section 5.1.4]{leytonbrown2022matching}.
Making such systems more powerful and more robust would require advanced NLP techniques. First, conflicts arising from supervisory relationships, co-holding grants, working at the same institution, being students of the same supervisor, etc. are hard to infer, as there exists no  single, comprehensive and reliable source of such data. A second, interlocking problem is the difficulty of name disambiguation. There exist various web services\footnote{Besides DBLP and ORCID, further examples include National Science Foundation (\url{https://nsf.gov}), and academic genealogy tree databases such as \url{https://genealogy.math.ndsu.nodak.edu} and \url{https://academictree.org}.} that provide information on co-authorship, joint-grant, and academic advising relationships. However, mapping information from such external databases to the reviewer and author database in a conference is a challenge since the same natural person may be associated with different name variations, multiple and changing affiliations, etc.~\cite{Amado_2023_registry}. Development of better natural language processing techniques to address this important issue can help improve the checking of conflicts in peer review as well as various other processes like correct attribution of citations, etc. Finally, it may be desirable to add constraints between reviewers who are in direct competition with authors' work---though this issue must clearly be approached with care, as competitors can also be the most knowledgeable reviewers.

\subsubsection{Reducing incentives for strategic behavior}\label{sec:colrings} The matching problem can be further adjusted to reduce the effectiveness of strategic behaviors such as collusion rings (Section \ref{sec:before_matching_bidding}). The AAAI-2021 conference addressed this through various constraints, based on the principle that being overly aggressive about forbidding potentially problematic matches (false positives) was better than allowing collusive behavior to succeed (false negatives), particularly given the large number of potential reviewers for each paper \cite{leytonbrown2022matching}. Specifically, this conference penalized matches in which multiple reviewers came from the same geographic area, in which any pair of reviewers had coauthored papers together (even if neither was in conflict with the submission), and in which any pair of reviewers had both bid positively on each other's papers and were both assigned these papers to review (because such ``2-cycles'' create opportunities for reciprocity). Other AI/ML conferences have further addressed incentives for strategic behavior by adding randomness to reviewer--paper matching scores to make the formation of collusion rings more difficult \cite{jecmen2020manipulation}. Yet, using only quantitative data like bidding, it is hard to detect collusion rings~\cite{jecmen2022tradeoffs,jecmen2024detection}---and combining this data with NLP analysis bears promise. For example, NLP can be used to re-check the expertise of the assigned reviewer and spot the outliers who may have manipulated bidding. Alternatively, NLP can be used to analyze whether a reviewer engaged in an unusual amount of discussion (Section \ref{sec:during_discussion}) to get a suspected COI paper accepted. The potential of NLP for alleviating the problem is yet to be explored.

\subsubsection{Prioritizing good reviewers} As conferences grow, curating sets of reviewers becomes increasingly difficult for the program chairs. Leveraging data about reviewers' past performance can help assemble reviewer sets by giving well-performing reviewers priority in subsequent matches. The most straightforward approach restricts itself to less sensitive metadata about reviewing: whether past reviews were submitted on time, how long reviews were, whether reviewers read author rebuttals and participated in discussions, etc. However, even using review metadata may create privacy risks, as prior work has shown that information about timing of reviews alone can aid in de-anonymizing the reviewers~\cite{goldberg_timing_privacy}. Natural language processing bears great potential in helping to determine review quality automatically, with caveats as detailed in Section \ref{sec:resource_eval_eval}. Yet, re-using the data from \emph{prior} conferences to determine a reviewer's \emph{past} performance by the current program chairs meets even greater confidentiality risks, and calls for novel applications of privacy-preserving techniques such as anonymization and differential privacy \mbox{(Section \ref{sec:ethical_legal})}.

\subsubsection{Manual assignment support} At many AI conferences automated reviewer--paper assignment is only the first step, followed by a manual adjustment of the results. This manual adjustment can be a laborious process that can benefit from NLP assistance. For example, NLP could be used to help identify areas of expertise that are required to thoroughly evaluate the submission, but are not offered by any of the currently assigned reviewers. This assistance scenario would be particularly valuable for emerging research areas, for which there are few experts, and for interdisciplinary submissions that can benefit from reviewing expertise outside of the default reviewer pool. For example, at ACL-2023, keyword frequency analysis was used to identify the papers with `rare' topics, languages or contribution types, and the senior meta-reviewers were asked to consider adjusting the assignments for these papers \cite{rogers-etal-2023-report}. Ideally, NLP-based tools would aid in identifying such papers more precisely, and could further be expanded to suggest appropriate reviewers and provide PCs with a simple way of inviting these reviewers, incl. explaining why their specific expertise is needed. Such NLP assistance could also be of use for reviewing systems where all reviewer assignments are manual, as done in many journals. 

\section{Assistance During Review}
\label{sec:during}

\begin{figure}[t]
\centering
\includegraphics[width=\textwidth]{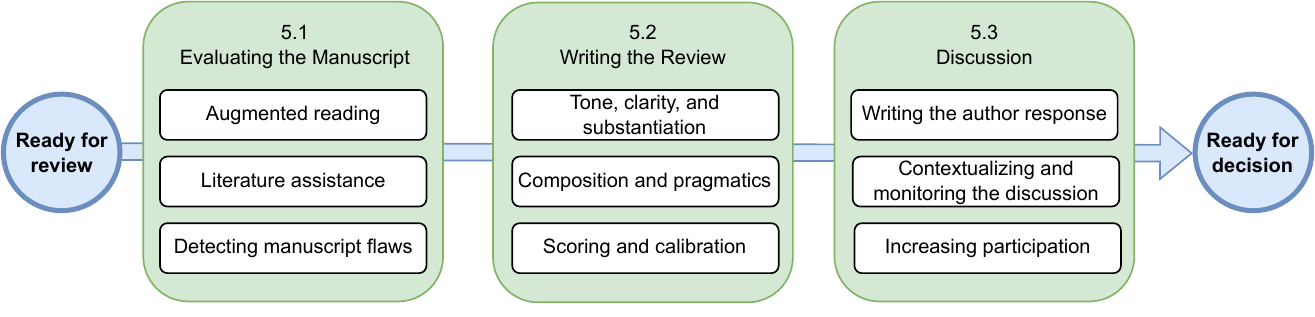}
\caption{Areas of assistance during peer review.}\label{fig:during_review}
\end{figure}

Once the submissions are distributed among reviewers, the reviewing begins. In this section we continue our discussion by outlining core challenges and opportunities for NLP assistance during manuscript evaluation, review-writing and the subsequent discussion between authors, reviewers, and meta-reviewers (Figure~\ref{fig:during_review}).

\subsection{Evaluating the Manuscript}
\label{sec:during_eval}

The review phase starts with the reviewers reading and evaluating the manuscript at hand. Reading for peer review is a hard expert task that often requires reviewers to build a deep understanding of the work and draw upon their background knowledge and analytical skills. The goal of NLP assistance at this stage is to reduce reviewer effort while promoting grounded and thorough assessment.

\subsubsection{Augmented reading} Recent years have seen substantial advances in developing AI-powered reading interfaces for scientific manuscripts. Such tools can be used to support peer review. For example, SCIM~\cite{fok2023scim} provides multi-class highlights on the manuscript text that can guide the reviewer through the general topic and important aspects of the work, such as contributions, experimental setup and findings. ScholarPhi \cite{head2022math} augments the reading application with symbol definitions for parts of mathematical formulae. Guided reviewing interfaces are another subclass of reading assistance for review: since novice reviewers would be unfamiliar with the process of writing a review, a tool that guides them through the process could improve review quality \cite{stelmakh2020novice}. For example, the recently introduced ReviewFlow \cite{sun2024reviewflow} directly approaches the problem of AI-based scaffolding to support novice peer reviewers. General-purpose frameworks for developing AI-augmented reading applications \cite{zyska-etal-2023-care, lo-etal-2023-papermage} facilitate further development of guided reading interfaces for peer review, along with question-answering tools to help manuscript navigation.

\subsubsection{Literature assistance} Reviewing requires in-depth understanding of prior work. Due to the fast pace and increasing topical diversity of modern research, such in-depth understanding might be lacking even for experienced reviewers. NLP has potential to assist literature work in several ways. Citation recommendation tools \cite{bhagavatula2018content} can be integrated into the reviewing process to help reviewers detect relevant work not mentioned in the manuscript, potentially guiding the judgement of novelty of the work \cite{wang2024scimon} and helping address the common problem of unjustified lack-of-novelty criticisms \cite{rogers-augenstein-2020-improve}. Such assistance would help reviewers suggest additional comparative discussions when needed, in the light of conflicting results in related work or missing baselines. Systems for citation recommendation and fact-checking \cite{wadden-etal-2022-scifact} may be adapted to help the reviewer assess whether the references that are present in the manuscript are cited appropriately, detecting factually wrong claims and misrepresented prior results, and potentially guiding the judgement of the work's soundness and motivation. As part of assessing the paper, reviewers often use search engines to find related literature---and might encounter public versions of the manuscript under review, which can inadvertently expose the authors' identities and affect the evaluation~\cite{rastogi2022arxiv}. Custom search engines that hide the manuscript under review from the search results obtained as part of the reviewing process could help mitigate this problem.

\subsubsection{Detecting manuscript flaws} Finally, NLP has the potential to help reviewers with more in-depth analysis of the manuscript. Recent work has shown that generative AI models could be used to identify errors such as mathematical mistakes, conceptual fallacies, or problems in experimental design~\cite{nuijten2020statcheck,liu2023reviewergpt}, but might overemphasize certain aspects such as implications of the research and adding more experiments, compared to humans \cite{liang2023can}. Multimodal applications of NLP to analyze the correspondence between numerical results (figures and tables) and their textual interpretation in the manuscript are another promising research direction \cite{blecher2023nougat}. While it is questionable whether the process of manuscript assessment itself can or should be fully automated, it might be possible to develop tools that nudge reviewers to consider typical pitfalls. For example, NLP assistance can be used to check if the manuscript complies with specific experimental standards of a research community, such as reporting of computational costs, statistical testing, study participant details, etc.---as outlined in reproducibility checklists widely employed in the NLP and AI community \cite{DodgeGururanganEtAl_2019_Show_Your_Work_Improved_Reporting_of_Experimental_Results}. 

\subsection{Writing the Review}
\label{sec:during_writing}

During paper evaluation, reviewers often take notes and draft the initial review. This draft then needs to be transformed into an official document addressed to the authors, meta-reviewers and program chairs, that constructively evaluates the paper, informs the meta-reviewers about the benefits and flaws of the submission, and instructs the authors on potential improvements. Low-quality reviews are frustrating and commonplace, indicating the potential for review writing assistance. The goal of NLP at this stage is to help the reviewer craft a useful review report and mitigate common issues prior to submitting it into the system. The issues detected at this stage can be taken into account by the reviewer for self-improvement, or by the meta-reviewers in case the issue persists (\cref{sec:after_meta}). 
Feedback to reviewers could be combined with concrete consequences after repeated violations of the established reviewing standards. Effective NLP tools to evaluate the quality of reviews may be directly useful in building incentives that encourage reviewers to provide high-quality work~\cite{goldberg2023peer}.

\subsubsection{Tone, clarity, and substantiation}
\label{sec:during_writing_tone} Reviewer anonymity aims to elicit objective reviews and protect the reviewers from potential backlash---yet anonymity also reduces accountability on the reviewer side. As a result, rude, vague and unsubstantiated reviews are not uncommon. Computational approaches to politeness analysis \cite{politepeer, 10.1145/3654660} can help mitigate communication issues. To address vagueness in peer reviews, specificity analysis \cite{Li_Nenkova_2015} can be deployed; \citet{10.1145/3529372.3533300} investigate hedging and uncertainty expressions in peer review. To improve review substantiation, claims that are inconsistent with the content of the paper or not supported by related work may be detected \cite{guo-etal-2023-automatic}. Matching reviewer comments to paper content can help detect the overall lack of anchoring to the submission text, detect misaligned comments, and point reviewers to specific sentences or passages in the paper with contrary evidence \cite{10.1162/coli_a_00455, d2023aries}. Another connection may be drawn between review substantiation and a broad line of work on scientific fact checking \cite{glockner2024ambifc}, where claims made by reviewers would need to be validated against wider bodies of scientific knowledge.

A special case of substantiation analysis pertains to heuristics and strategic behavior. Peer review is a hard task prone to heuristics, such as \textit{``results are not surprising''}, \textit{``results do not surpass state of the art''}, \textit{``the method is too simple''}, \textit{``the authors should have done [X] instead''}, and others.\footnote{See \url{https://aclrollingreview.org/reviewertutorial} for more examples.} Due to the competitive environment, reviewers might behave strategically, for example, by enforcing citations of their own work or by writing ``torpedo reviews'' aimed to scuttle certain topics or competing groups. Heuristics and strategic behavior might be observed in review text, and thus could be automatically detected by NLP methods, helping the reviewers check the thoroughness and impartiality of their argumentation. However, the aforementioned issues count as such \textit{only if} a particular evaluation or statement is \textit{not substantiated} or faces contrary evidence. For example, simply rejecting a work for not delivering a state of the art score is a heuristic. Yet, if results do not surpass state of the art \textit{while the manuscript incorrectly claims otherwise}, it is a legitimate concern. Similarly, a reviewer suggesting highly relevant work, which happens to be their own, is not necessarily behaving strategically. The problem of detecting heuristics and strategic behavior thus goes beyond surface-level analysis, and constitutes an exciting direction for future NLP research. Finally, as with manuscripts, plagiarized and fake reviews are also an issue~\citep{piniewski2024emerging} that NLP tools can help detect and prevent. 

\subsubsection{Composition and pragmatics}\label{sec:during_writing_prag} Peer review is an argumentative text that pursues a range of communicative goals. Yet, unlike research papers that generally follow an established discourse structure, the standards of peer review writing are less codified. NLP assistance can be thus deployed to enforce the adherence of peer reviews to the composition standards accepted in a given community. One potential line of work contributing to this is the discourse analysis of peer reviews: pragmatic tagging \cite{10.1162/coli_a_00455, dycke-etal-2023-overview} aims to label each review sentence according to its role (such as strength, weakness, todo or summary), while argument-mining approaches \cite{hua-etal-2019-argument} discover the argumentative structure of reviews, splitting them into facts, requests, and evaluations. This can be coupled with the argumentative analysis of the submission itself~\citep{lauscher-etal-2018-investigating, lauscher-etal-2022-multicite}. 

A traditional way to enforce composition of peer review reports is to deploy structured and semi-structured reviewing forms---which has been shown to improve inter-rater agreement of decision recommendations \cite{malivcki2024structured}. Structured forms open new opportunities for NLP assistance in peer review. On one hand, NLP can be used to help reviewers pre-fill structured reviewing forms based on their drafts and reviewing notes, and to detect inconsistencies (e.g., a substantial weakness listed as a question). On the other hand, data from structured reviews can be used to improve review quality in non-structured reviewing---for example,  \citet{dycke-etal-2023-overview}  use structured review forms as weak supervision for analyzing essay-style reviews. This analysis could be applied to provide feedback to reviewers during writing and be particularly useful for novice reviewers.

\subsubsection{Scoring and calibration} 
\label{sec:during_writing_calibration}

To facilitate decision making at the later stages of the process, reviewers are often asked to accompany their reports with numerical scores. An overall score and a confidence score are commonly used; some conferences employ additional ``criteria scores'', like novelty, soundness, impact, or excitement. While convenient, scoring meets multiple challenges. Given the high effort associated with reviewing, each reviewer normally evaluates only a small subset of submissions. This naturally leads to calibration issues. One kind of miscalibration arises when reviewers have different interpretations of the scores or thresholds for acceptance, with some reviewers being stricter, some lenient, and so on~\cite{wang2018your}. Several algorithms have been developed to address this issue~\cite{roos2012statistical, ge13bias} but have not performed well in practice (see~\citealp[Section 5]{shah2022surveyextended} for more details on this miscalibration). These algorithms rely solely on the numerical scores provided by reviewers, and can be complemented by NLP analysis. Score prediction \cite[and others]{kang-etal-2018-dataset, li-etal-2020-multi-task} can be used to detect inconsistencies between review text and the overall score, and confidence score prediction can similarly be used to detect both over-confident and under-confident reviews. This information can help the reviewers adjust their scores prior to submitting the review, and can be especially useful for junior reviewers and researchers who review for multiple conferences which employ different scoring schemata and score semantics. At an overarching level, NLP tools that can access the text of reviews across the conference (or even multiple conferences) can help calibrate reviews across the entire reviewer pool. Another calibration issue occurs when reviewers employ different ways of combining individual criteria (such as novelty, clarity, etc.) into their overall recommendation score---often referred to as ``commensuration bias''~\cite{lee2015commensuration}. Existing solutions to commensuration bias such as ~\cite{noothigattu2018choosing} are limited due to their reliance on numeric scores alone, and NLP can substantially contribute to this line of research by helping extract individual preferences from reviewers' aggregate review reports.
\subsection{Discussion}
\label{sec:during_discussion}

After the initial round of reviews is completed, peer review enters the discussion phase. Here, the authors communicate with their reviewers for the first time, and can address reviewers' questions and concerns, and ensure that the key contributions of the paper are properly understood. This is followed by a discussion among the reviewers---often prompted by the meta-reviewer. Here, reviewers can edit their review reports and adjust the scores to reflect their final assessment of the manuscript. This discussion creates a complex dialogue grounded in the context of the paper, reviews, and the conversation so far, and offers ample space for NLP assistance. The goal at this stage is to ensure that the discussion is efficient and effective, and results in a more objective evaluation of the submission, which later informs the meta-review (Section~\ref{sec:after_meta}).

\subsubsection{Writing the author response} The authors are presented with a set of reviews discussing the paper and given a short period of time to respond. Like reviews, author responses can be done in ways that are more or less productive, and similarly to review writing support (\cref{sec:during_writing}), authors can be provided with feedback about the composition and tone of their author responses, including argument mining, pragmatic tagging, politeness, and specificity analysis \cite{gao-etal-2019-rebuttal}. Recent work has explored helping authors argue for their submission via attitude- and theme-guided rebuttal generation \cite{purkayastha-etal-2023-exploring}. NLP tools can further help the authors extract key questions and concerns from the peer reviews, e.g., by extracting individual action items, aggregating points of concern, and helping the authors devise a plan for the author response, as outlined by the widely used tutorial on rebuttal-writing in the ML community.\footnote{See \url{https://deviparikh.medium.com/how-we-write-rebuttals-dc84742fece1}.}

\subsubsection{Contextualizing and monitoring the discussion}\label{sec:during_discussion_context} Due to the author response period, the discussion among reviewers can take place several weeks after the initial reviewing. Because of this time lag reviewers may struggle to interpret parts of the discussion pertaining to the submitted manuscript, other reviews, and author responses. 
To help guide the discussion, cross-document analysis can be applied to identify and give the reviewers clear pointers to the parts of the manuscript under discussion \cite{10.1162/coli_a_00455, d2023aries}. NLP assistance can provide direct mappings between the author responses and review reports \cite{cheng-etal-2020-ape}, and help analyze the discourse structure of the author responses, e.g., by highlighting parts of the response that address reviewers' criticisms and reporting on the changes in the submission text in response to the review \cite{kennard-etal-2022-disapere}.

The organization of the discussion phase depends on the conference management system and the venue. For example, in the author response process at a typical AI conference, all participants need to engage in the discussion and track updates that apply to them. Most current CMS can either send notifications for all updates, which can be overwhelming, notifications of explicit replies, which may be insufficient, or no notifications, which puts the onus on the participants to track discussion. Basic natural language processing could assist by detecting updates to the conversation that concern each participant and notifying them accordingly. Similarly, NLP tools can be used to notify meta-reviewers in cases their intervention is mandatory---such as open conflict between reviewers and authors, related to recent work on dispute tactics in Wikipedia \cite{de-kock-vlachos-2022-disagree}.

The utility of the discussion phase is often debated: while some studies find that the discussion phase did little to alter reviewers' opinion~\cite{daumesome, shah2018design}, others reveal favorable outcomes~\cite{parno2017report, frachtenberg2020survey}.
Past studies also hypothesize and experiment with \emph{anchoring} and \emph{herding} effects among reviewers, 
wherein a reviewer might be 
(disproportionally) 
influenced by other reviewers~\cite{teplitskiy2019social} or might follow the reviewer who initiates the discussion~\cite{stelmakh2023large}. 
Often such anchoring and herding effects are undesirable, and NLP techniques may assist 
in tracking and flagging the influence of reviewers towards each other by 
analyzing the discussion and the subsequent changes in the review reports.

\subsubsection{Increasing participation}\label{sec:during_discussion_enforce} Reviewing takes substantial time and effort, and is rarely the reviewers' primary professional occupation. Low reviewer participation in the discussion phase is commonplace~\cite[Section 3.5]{shah2018design}. Two possible approaches to promote discussion among reviewers are generic reminders---which are often ineffective and ignored---and personalized reminders---which are generally known to be more effective \cite{vervloet2012effectiveness}, but require effort by the meta-reviewers. Given the capabilities of modern generative AI for text, NLP tools may assist meta-reviewers in nudging reviewers to further engage in the process, from customizing and pre-filling reminder templates to generating personalized reminder emails that can be finalized by the meta-reviewers and sent to participants. A deeper level of NLP assistance can be deployed to detect reviewer statements and author responses that go unanswered or are not sufficiently followed through. A prototypical and familiar case is when a reviewer points out a concern which leads them to assign the work a low overall score. In their response, the authors fully address the concern. In the discussion phase, reviewer thanks the authors for their response, but \textit{does not adjust the score}. As result, the authors notify the meta-reviewer about potential unfair reviewing, or, alternatively, simply accept their fate and hope that the issue is noticed during meta-review. Automatically detecting such cases during the discussion when the reviewers are still present and engaged has a potential to improve the quality of the process and reduce meta-reviewers' workload at later stages of peer review.

\section{Assistance After Review}
\label{sec:after}

\begin{figure}[t]
\centering
\includegraphics[width=\textwidth]{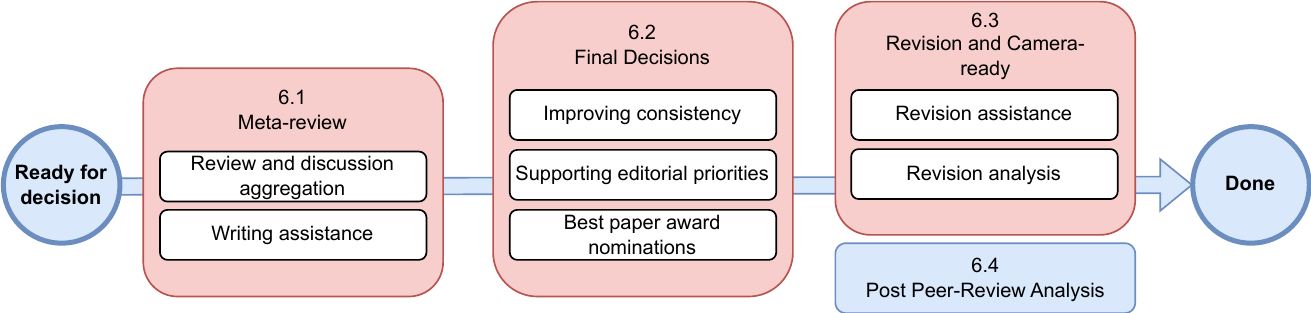}
\caption{Areas of assistance after peer review.}\label{fig:after_review}
\end{figure}

Once reviewing is finished, the reviewing campaign enters its final phase. Following the terminology introduced in \cref{sec:define}, the review reports from the reviewers are aggregated by the meta-reviewers (area chairs); these are later used by senior meta-reviewers (senior area chairs) to make accept/reject recommendations, which inform the program chairs (PC) in the final decision-making phase. We now discuss the core challenges and opportunities in applying NLP to support meta-review writing, decision making, camera-ready revision preparation, and post-analysis of the reviewing campaign (Figure \ref{fig:after_review}).

\subsection{Meta-review}
\label{sec:after_meta}

Meta-reviews are the first step in the decision-making process. A meta-reviewer's duty is to aggregate input from individual reviews, detect low-quality reviews, find emergency reviewers if needed, and oftentimes to read the manuscript themselves and provide their own grounded opinion. The result of this work is a meta-review---a brief report to the PCs that summarizes individual peer reviews and discussions, often including a meta-reviewer's own evaluation. Meta-review also summarizes feedback and required changes for the authors and provides an initial acceptance recommendation. Preparing a high-quality meta-review is hard: the meta-reviewer must read each review, rebuttal, and discussion thread, aggregate the information, take a stance on controversial points, and assess whether the authors' responses sufficiently address the reviewers' concerns. NLP assistance at this stage aims to facilitate this complex multi-document analysis task. 

\subsubsection{Review and discussion aggregation} 
The meta-reviewer's task of aggregating information from the prior peer-reviewing data meets three core challenges: the amount of information that needs to be processed, the varying quality of this information, and the inherently argumentative nature of the underlying texts. To help meta-reviewers navigate and interpret the reviews, same NLP approaches that can help reviewers compose their reports (Section \ref{sec:during_writing_prag}) can be adopted to separate peer reviews into statements and categorize these statements according to their argumentative or pragmatic function. This way, a meta-reviewer can quickly aggregate information from different peer reviews, e.g., get an overview of the manuscripts' strengths, weaknesses, and improvements suggested by multiple reviewers. We note that peer reviews afford analysis across multiple dimensions, which can all support meta-reviewers in their work. In particular, aspect-based analysis of peer reviews \cite{Chakraborty_2020, kennard-etal-2022-disapere} can help identify aspects of the manuscript that were insufficiently or vaguely covered by the reviewers. For example, lack of evaluation in the ``technical soundness'' aspect can prompt the meta-reviewer to investigate the manuscript more carefully, and ultimately help prevent accidental acceptance of technically flawed work. We note that existing works on aspect-based analysis of peer reviews operate with fairly coarse-grained aspect labels. A finer granularity dependent on the contribution type might bring further benefits: for example, a dataset paper would not require rigorous discussion of the experimental setup, but would instead prompt more detailed scrutiny into questions such as licensing, participant recruiting and compensation, guidelines, etc. To help meta-reviewers navigate the discussion around the manuscript, approaches building upon discussion assistance (Section \ref{sec:during_discussion_context}) can be re-used at this stage, e.g., by providing explicit links between the manuscript, the reviews, author's responses and discussions.

A further level of assistance can be provided via meta-review generation---automatic summarization of peer reviews, discussions and author responses. Such summaries can include a list of contributions, weaknesses, and a summary of the discussion, ideally taking into account reviewer confidence and expertise. This task is challenging due to the need to process diverse, interconnected texts which contain differing, often antagonistic opinions. Authors want their manuscript to be accepted: they will argue in favor of their paper, possibly trying to minimize issues identified by the reviewers or to provide easy solutions to reviewers' concerns. Reviewers want the manuscript critically assessed: they will focus on the perceived weaknesses and might disagree with each other. Meta-review generation is related to aspect-based sentiment analysis and multi-document summarization, and has recently received increased attention. \citet{zeng2023meta} cast the task as scientific opinion summarization. They propose a checklist-based approach to prompt an LLM to write summaries that fulfill meta-review requirements of self-consistency, faithfulness, and active engagement. \citet{li-etal-2023-summarizing} explore a summarization model that incorporates inter-document relationships between reviews, author responses, and discussions, finding that the model struggles with recognizing and resolving conflicts. \citet{santu2024prompting} conduct in-depth experiments with prompting LLMs to perform the task. \citet{shen-etal-2022-mred} and \citet{9651825} explore meta-review generation in a controlled generation setting conditioned on the meta-review score. These initial approaches bear promise that NLP will eventually be able to support meta-reviewers in navigating the complex reviewing discourse. Yet, given the importance of meta-reviews for decision making, we caution against the use of NLP to fully automatically compose meta-review reports and take decisions on meta-reviewers' behalf.

\subsubsection{Writing assistance} While meta-reviewers are typically more experienced than reviewer pool average, they work under stricter time constraints and experience higher workload than regular reviewers. Many aspects of review-writing also apply for meta-review, and meta-reviewers can benefit from similar NLP assistance. Similar to review reports, tools can help ensure that the meta-reviews are clear and well-substantiated (Section \ref{sec:during_writing_tone}) and include key information necessary for decision making by the PCs such as manuscript strengths, weaknesses and required improvements (Section \ref{sec:during_writing_prag}). NLP assistance can further help calibrate meta-review scores and acceptance recommendations (Section \ref{sec:during_writing_calibration}). It is important to note, however, that meta-reviews and review reports serve different purposes, and many of the reviewing assistance techniques would need adjustment to help meta-reviewers. While the main purpose of a review report is to evaluate the work as thoroughly as possible, a meta-review aims to facilitate \textit{decision making} by the program chairs, and the most helpful meta-reviews are the ones where the meta-reviewer unequivocally takes a stance in favor or against a paper. Approaches to hedging and uncertainty detection \cite{10.1145/3529372.3533300} thus gain in importance in meta-review writing assistance. Similarly, while reviewers might give authors improvement suggestions which the authors are free to adopt or ignore, meta-reviewers can \textit{condition} manuscript acceptance on fulfilling some new criteria, marking the need for not only detecting, but also prioritizing required changes. While reviews must be substantiated based on the manuscript and related work, meta-reviews must additionally take into account the content of author responses and discussions, etc.

Overall, the high stakes at this stage of peer review motivate the development of assistance tools to ensure the highest possible quality of meta-reviewing. At the same time, given the high stakes, increased care needs to be taken as not to bias the meta-reviewers and distort their decisions. Thus, full automation of this stage of peer review is ill-advised at the current state of NLP technology, and possibly on general ethical grounds (Section \ref{sec:ethical_legal}).

\subsection{Final Decisions}
\label{sec:after_decision}

Once meta-reviews are completed, program chairs (PCs) are presented with meta-reviews for each manuscript, the original reviews, author responses and discussion, and the paper itself, and make the final accept/reject decision for each paper. Peer review is an imperfect process, and falsely accepted and rejected submissions are not uncommon. Given access to the fullest information about the peer-reviewing process up to this stage, PCs can use NLP assistance to address issues with the reviewing process and detect deficiencies in the meta-reviews. Yet, similar to our stance on meta-review assistance, we advice against a fully automated approach to assessing the value of scientific papers. The goal of NLP assistance at this stage is to provide additional input and aggregate existing information to increase it's utility for decision making, and never to replace expert human judgement.

\subsubsection{Improving consistency} Reviewed manuscripts typically fall into three categories: easy accept papers, easy reject papers and borderline papers that require careful consideration. At major AI conferences, the borderline paper pool is very large, from hundreds to thousands of papers. Automatic tools can be developed to help PCs in ranking borderline papers according to a combination of PC-defined criteria. NLP tools can assist decision making by grouping borderline papers in terms of their topic, contribution type, as well as strengths and weaknesses as identified by the reviewers and meta-reviewers. A dedicated tool can then keep track of decisions made for similar papers across different topics or generate clusters of decisions to facilitate comparison and minimize cases where a paper is accepted, and another comparable paper is rejected. For example, in clusters from the same track, PCs might want to compare the assertiveness of the meta-review comments, and rank papers based on this aspect: prior work on detecting uncertainty and hesitation \cite{10.1145/3529372.3533300} could be extended for these cases. 

Another potential direction for NLP assistance at this stage is to facilitate \textit{pairwise} paper and review comparison in case of ties, e.g., by aggregating individual contributions of the papers based on review reports and meta-reviews. Often the decision-making process requires a calibration step prior to ranking papers on any aspect of the meta-reviews. Similar to reviewer score calibration (Section \ref{sec:during_writing_calibration}), NLP tools can be used to compare meta-review scores and comments, and allow a possible scaling of meta-reviews that can reduce unintended and unfair assessments. At a minimum, PCs can be given a type of ``strictness score' for the the meta-reviewers that can be used by PCs to better calibrate and rank papers.

Finally, NLP tools can assist PCs in detecting cases where the meta-review appears to be taking a significantly different stance on the overall assessment of the paper compared to the reviewers. Cases found to have large disagreements in this sense might point to situations that need careful consideration as they may indicate unethical meta-reviewing. For example, when a meta-reviewer defends a paper, but reviewers have found the that paper has major flaws that cannot be addressed by the camera-ready deadline, PCs need to intervene and ensure that the meta-reviewer is not colluded with authors; in the case where the meta-review is overly negative and the reviews are not, PCs need to evaluate whether the meta-reviewer is biased or is acting in an unprofessional manner. 

\subsubsection{Supporting editorial priorities} While manuscript evaluation by the (meta-)reviewers often has clearly defined criteria and is supported by extensive guidelines, final decisions by PCs are typically less structured for the larger set of borderline papers. Some PCs might want to rank papers on certain topics higher than papers on other topics in the same borderline batch, based on their editorial priorities for a given venue. Current CMS and peer review tools allow specifying formulae based on numerical scores in the reviews---yet this is insufficient for the PC's decision-making process, which often involves the text of the reviews and meta-reviews, the discussion, etc. Creating an automated tool that can adjust on the fly to the ranking specified by the PCs would be a potentially useful application to explore. As an example, PCs could explicitly define their ranking criteria in natural language: ``\textit{rank all papers in the borderline group based on the largest disagreement in overall assessment, and then based on their relevance to the special theme}''. This definition could then serve as input to NLP assistance systems that would make use of the combined information in the manuscript text and review reports to select and prioritize the relevant subset of submissions. 

\subsubsection{Best paper award nominations}

The technical programs in conferences typically include (best) paper awards. Given the large submission numbers, candidate submissions are usually determined either by explicitly asking reviewers to nominate the papers they are reviewing, or by considering manuscripts that have received a very high score among reviewers.\footnote{See \url{https://www.aclweb.org/adminwiki/index.php/ACL_Conference_Awards_Policy} or \citet[Section 2.3]{shah2018design}.} Both approaches face challenges. Reviewer's nominations are often scarce: for example, at the ACL-2023 conference, almost all submissions considered by the best paper committee had only one reviewer nomination. Because of this, many equally deserving submissions get overlooked because they never get nominated. Relying on the scores, in turn, is prone to calibration issues (Section \ref{sec:during_writing_calibration}). One approach to increasing fairness of best paper selection would be to augment existing mechanisms with automated tools that take review texts into account. Given a set of high-scoring or nominated submissions, such tools could detect papers that have received similar comments about their novelty, soundness and contributions. Approaches to detecting innovative work as is done by \citet{10.1073/pnas.1915378117} could be used here to identify papers that are unique and worthy of adding to the pool of best paper candidates. This automatically augmented pool of papers can be then passed on to the best paper award committee for ranking and selection of best papers. 

\subsection{Revision and Camera-ready}
\label{sec:after_revision}

Whether accepted or rejected, the manuscript is revised at least once as a result of the peer-reviewing process. If the manuscript is accepted, the authors must incorporate reviewer suggestions into the camera-ready version (\textit{``Please provide p-values for Table X''}), as well as follow up on the commitments made in the rebuttal (\textit{``We will add this to the limitations''}). Similarly, if the manuscript is rejected, the authors are expected to use the reviewers' feedback to make the manuscript stronger before resubmitting it to another venue. Revision is often accompanied by an amendment note where the authors describe the changes made, submitted along with the revised manuscript. NLP has the potential to assist the participants of the peer-reviewing process in two core ways: by helping authors revise the manuscript based on feedback, and by helping reviewers, meta-reviewers and program chairs analyze the resulting changes.

\subsubsection{Revision assistance} Ideally, the revision should incorporate reviewer and meta-reviewer feedback as well as the authors' commitments made during discussion. This can result in a large number of interconnected edits of different localization and complexity that need to be prioritized and executed while adhering to the formal requirements like page limitations. Recent works in NLP pave the path towards automatic edit localization by connecting reviews and rebuttals to the manuscript text \cite{10.1162/coli_a_00455, d2023aries}; suggesting and executing some of the revisions has been also explored and proven a challenging task \cite{d2023aries}. Once reliable NLP technology is available, user-facing applications that build upon it would need to overcome the variation in reviewing and publishing workflows between different research communities and venues, as well as technical challenges related to processing complex, richly formatted long documents (\cref{sec:resource_eval_data}). Besides, the space of potential manuscript revisions is very diverse, from correcting a single-word typo to including new experimental sections. Not every such edit operation can or should be automated. Edit prioritization and routing of the revisions between NLP assistants and human authors is an exciting avenue for future research. Automatic edit summary generation to help the authors compose amendment notes is yet another promising research direction.

\subsubsection{Revision analysis} The revision step raises new questions for reviewers, meta-reviewers and program chairs as well. Does the revision address all the points raised by the reviewers and follow up on the promises made during the rebuttal? Currently, nothing technically prevents the authors from entirely ignoring some of the reviewer suggestions once the manuscript is accepted. Does the revision feature new material that has not been reviewed and that significantly alters the substance of the manuscript? Again, at the moment, nothing stands in the way of this, and manually re-analyzing the changes for each submitted manuscript is similarly not feasible. These use cases motivate automatic analysis of manuscript revisions. While basic \texttt{diff}-like functionality is commonplace in conference management systems like OpenReview, more advanced processing of revisions has been approached for arXiv pre-prints \cite{jiang-etal-2022-arxivedits, du-etal-2022-understanding-iterative} and ACL publications \cite{mita2022automated}, and in the context of peer review \cite{10.1162/coli_a_00455, d2023aries}. Existing NLP approaches help align manuscript revisions on sentence and paragraph level, detect new and changed content, and can automatically label the edits by their intent, e.g., grammar, clarity or factual adjustment. These pilot studies can already support both manuscript-level analysis to aid the reviewers and meta-reviewers, and collection-level analysis that can provide general, high-level insights about the revision process across different research communities or intervention groups (\textit{``Does the new policy result in more thorough revisions?''}). Potential directions for follow-up work in this area include multimodal processing to incorporate tables, formulas and figures into the analysis; designing and evaluating user-facing applications to help reviewers and the editorial team to check the work; and meta-studies on the general revision behavior across research communities.

\subsection{Post Peer-Review Analysis}
\label{sec:after_post}

Peer review generates large and complex empirical data that can help us analyze and improve the process. Examples of such analysis include post-conference reports~\cite{shah2018design, rogers-etal-2023-report} that cover various aspects and changes to the review process for a given conference, or evaluations of interventions in the peer-review process, such as the reviewer training experiment of \citealp{stelmakh2020novice}. Post peer-review analysis typically amounts to reporting minimal summary statistics, that are presented at the conference and recorded in the front matter of the proceedings. Further analysis is labor-intensive and requires additional effort from the few participants with full access to the reviewing data (program chairs and technical staff), who are typically exhausted at the end of the reviewing campaign. This motivates the development of automated tools for post-peer review analysis.

NLP assistance could open up a range of avenues for analysis beyond the standard statistics compiled today.
First, it could help us learn more about our reviewing. We could identify reviewer trends and biases,
for example, identifying which requests from reviewers correlate with acceptance decisions.
Those requests may be legitimate, and sharing them may help future authors avoid common pitfalls, or they may be problematic, indicating areas where additional guidance is needed for reviewers. Aggregate analysis of reviews, discussions, meta-reviews and revisions can give further insights, e.g., on what forms of discussion influence the meta-review the most, and for what reasons people adjust their review reports. Initial results in NLP-supported aggregated analysis of reviewing and editing behavior in peer review show great promise: among others, \citet{10.1162/coli_a_00455} investigate the connections between review comments and manuscript edits, while \citet{hua-etal-2019-argument} use argumentative structure to compare review composition in different computer science venues and investigate the relationships between review argumentation and scores. Yet, existing results are preliminary and motivate future work. 

Post peer-review analysis opens new opportunities to evaluate the reviewing process as a whole. Is our peer review able to accurately predict the impact of the work and reliably evaluate its soundness? To answer this, one might elicit impact assessments from the reviewers and relate them to some metric of future impact of the work. One simple---yet imperfect---metric of impact is the number of citations that a manuscript accumulates \cite{Plank2019CiteTrackedAL}, potentially refined by citation type or class, such as background citations and method citations \cite{teufel-etal-2006-automatic,
aburaed-etal-2017-sentence,jurgens-etal-2018-measuring,pride-knoth-citation20,lauscher-etal-2022-multicite}. The insights from this kind of automatic analysis would allow us to evaluate the effects of reviewing policies, gain a better empirical understanding of how peer review works, and hopefully make the process as a whole more transparent and trustworthy. 

Finally, post-review analysis can help us learn more about researchers. Conducting reviewer surveys is commonplace at AI conferences \cite{emnlpreviewersurvey2017, rastogi2022arxiv, shah2023role}. The results of these surveys can impact policies at future conferences. Typically, the surveys include structured and free-text fields. While structured fields can be easily aggregated and analyzed, free text and open-ended questions require careful reading and aggregation by the conference organizers. NLP tools for the analysis of free-text responses (e.g., by clustering or extracting common topics) could help reveal new trends and patterns in peer review and efficiently elicit feedback on new policies and common issues encountered by the reviewers. Assuming that the peer-reviewing data is made available (Section \ref{sec:resource_eval_data}), one could study trends in the scientific process over time, and compare reviewing practices across communities and disciplines. What questions do reviewers at NLP conferences ask today compared to five years ago, and how does that reflect the shifting norms in the field? Are these questions the same at other AI conferences, or are there differences? How does reviewing at AI conferences differ from human-computer interaction or computational social science venues? As research becomes increasingly multi-disciplinary, these insights might be highly valuable for scientists both while evaluating unfamiliar contribution types, and while submitting their own work to new communities.

\section{Data, Privacy, and Copyright}
\label{sec:resource_eval_data}

Developing NLP assistance for peer review requires data. Several unique features of peer-reviewing data set it apart from more traditional and well-covered application domains in NLP like Wikipedia, newswire, social media, and even the closely related domain of scientific publications. In the following we focus on the data that directly results from the peer-reviewing process, using AI conferences as a representative example for conference-style peer review. While we touch upon legal aspects of peer-reviewing data collection, we stress that this discussion \textbf{does not} constitute legal advice. We particularly recommend consulting legal experts when collecting reviewing data from a new venue or community.

\subsection{Data Composition} 

\begin{figure} 
\centering
\includegraphics[width=0.85\textwidth]{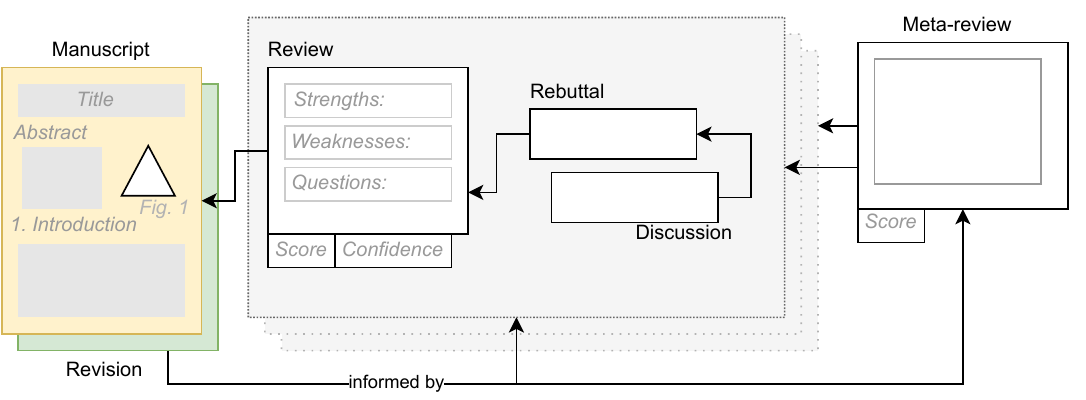}
\caption{Peer-reviewing data includes many complex document types that are semantically and structurally interrelated.}\label{fig:data_closeup} 
\end{figure}

Peer-reviewing data is complex and diverse (Figure~\ref{fig:data_closeup}). It covers a range of document types: the manuscript itself, its review reports, rebuttals, discussion threads, meta-reviews, and---potentially---a camera-ready revision. Many of these documents are \emph{multimodal} in a broad sense, combining written language with categorical metadata, structure (e.g., manuscript layout), non-linguistic elements (e.g., figures, tables, formulae, formatting), and numerical scores (e.g., overall and confidence scores). In addition, peer-reviewing data is highly \emph{interconnected}, where later documents are \emph{informed by} the earlier documents: the review does not exist in isolation, but discusses and actively refers to the manuscript; rebuttals are not standalone texts but act as replies to the reviews, and meta-reviews emerge as weighted summaries of individual (potentially revised) reviews and the subsequent discussion. 

This complexity brings new supervision and evaluation signal for NLP systems that would be hard to obtain otherwise. For example, if review reports are associated with numerical scores, one can directly use this information to learn to predict score from text, which in turn can aid with score calibration (Section \ref{sec:during_writing_calibration}) and outlier detection. At the same time, the complexity and inter-connectedness of the reviewing data bring new challenges. NLP has traditionally focused on the processing of isolated texts reduced to plain written language. While structured representations for scientific texts are available \cite{lo-etal-2020-s2orc, 10.1162/coli_a_00455, blecher2023nougat}, this information is rarely preserved when creating peer-reviewing data, limiting its further use; general-purpose formalisms for representing cross-document relations in peer review are equally scarce. The complexity of the data also affects task design, as it introduces new, potentially crucial contextual and multimodal information. Thus, when collecting peer-reviewing datasets, we deem it important to preserve as much information as possible, including non-textual elements of review forms, manuscripts in the original submission format, etc.

\subsection{Data Availability} 

Peer-reviewing data is abundant, yet \textit{open} peer-reviewing data is scarce, and NLP for peer review suffers from the lack of domain diversity. For a concrete example, the ACL-2023 conference alone generated over 12,000 reviews for over 4,500 full-text manuscripts \cite{rogers-etal-2023-report}, together with the accompanying meta-reviews, discussions, and revisions. Yet, apart from the camera-ready manuscripts for the accepted papers, none of this data is public. While some other communities and publishers make the reviews openly available e.g., F1000Research (\url{https://f1000research.com}), PeerJ (\url{https://peerj.com}), ICLR (\url{https://iclr.cc}), despite its growing popularity \cite{wolfram2020open}, open review is an exception rather than a rule. Three strategies for dealing with the lack of open peer-reviewing data emerge. First, one can develop NLP systems and conduct experiments using \textbf{closed data}---as it has been done in earlier meta-scientific studies of peer review \cite[etc.]{bornmann2011scientific,tomkins2017}. Alternatively, the data can be distributed under special conditions via a waiver---similar to the common practice in clinical NLP and other fields dealing with sensitive data. Yet, closed data is problematic in terms of reproducibility, makes follow-up research and deployment of derivative NLP systems challenging, and brings new administrative and technical challenges of redistributing the data and tracking its use. 

Second, one might resort to \textbf{open data} from few select venues and communities. Datasets based on open reviewing constitute the majority of peer-reviewing data available to NLP. Yet, challenges remain here as well. Since open reviewing data comes from only few communities, it might not be representative of the diverse reviewing practices in a new community or discipline. The sole fact that reviews are made publicly available (sometimes with reviewer identity disclosed, such as at F1000Research) might affect the tone and composition of the review reports, and some publishers only release peer reviews for accepted papers. Moreover, the data being openly accessible \emph{does not mean} that the data can be used for research purposes and openly redistributed. While some publishers are explicit about their licensing and terms of use (e.g., F1000Research), in other cases the legal status of the publicly available reviews and manuscripts is unclear. The third option for handling peer-reviewing data is \textbf{donation-based data collection}, where the data from a previously closed reviewing process is made openly available. The advantage of data collection on a donation basis is that it theoretically can be applied to any community and any venue. The disadvantage is that it takes time and effort, and needs to take into account the confidentiality, personal data regulations, and licensing.

\subsection{Considerations for Data Collection} 
Peer-reviewing data in a closed-review system is confidential. Making this data open via donation-based data collection requires \textbf{explicit permission} from the participants. Such permissions can take the form of terms of service and policy made known to the participants before the start of the reviewing campaign, with or without an option to opt out, like it is done at F1000Research and---since at least September 2023---at OpenReview.\footnote{\url{https://openreview.net/legal/terms}} Alternatively, the permission can be given by the participants on a case-by-case basis in an opt-in fashion. An additional challenge stems from the fact that peer-reviewing data is interconnected. In particular, any data associated with an unpublished manuscript (reviews, discussions) can leak information and results that can be misused by malicious parties (e.g., by scooping research ideas) and lead to resubmission bias, where knowledge about a paper's previous rejection at a conference can negatively influence subsequent reviews~\cite{stelmakh2020resubmissions}. Potential counter-measures include only collecting the data associated with accepted papers, or making the data public with a substantial time lag that would allow the authors to disseminate their work in the meantime, even if rejected. A direct consequence of all these strategies is that with each step the data becomes less representative of the full data of a reviewing campaign.\footnote{Which, in fact, can be rarely represented in full, as some variables like reviewer identities or their bids are very likely to never be made public.} One solution to this issue is to quantify the difference between closed full data and partial open data by using non-confidential aggregate statistics like score distribution, vocabulary overlap, etc., and report the comparative statistics between the open sample and the full population upon data release. 

Some of the peer-reviewing data relates to natural persons (authors, reviewers, meta-reviewers, program chairs), and thus constitutes \textbf{personal data under GDPR} (\url{https://gdpr.eu}), which regulates the use of personal data of EU subjects \emph{anywhere in the world} \cite{rogers2021just}, and is widely considered as a best-practice in personal data management overall. While GDPR is more permissive to research use than commercial use \cite{kamocki-etal-2018-data}, it still imposes requirements on the collection of reviewing data, including informed consent, a clear reflection on what exact data is collected and for what purpose, how the data is stored and managed, whether the data can be withdrawn and consent revoked, etc. 

Finally, peer-reviewing data constitutes intellectual property of its respective authors. Thus, in order to make peer-reviewing data openly available for follow-up research, it must be associated with a \textbf{license} that regulates its downstream use. Peer-reviewing data includes diverse document types authored by multiple parties, and many peer-reviewing data types fall into the grey zone of the common licensing practice. Some publishers make manuscripts openly available under a permissive license---but what about the initial submitted drafts and peer review reports? Who should be allowed to use the data and in which way? Who should be attributed as the data owner and who should stay confidential? Who should manage and distribute the licensed data, and how should a data repository be selected for secure long-term storage? These questions invite additional reflection and consideration while designing a data collection campaign; for a broader discussion see \cite{bender2018data,kamocki-witt-2020-privacy,rogers2021just}. A special case that deserves a mention is the use of third-party services to process peer-reviewing data, which became increasingly common with the advent of closed-source LLMs accessible via APIs. We stress that \textbf{processing peer-reviewing data off-site} requires careful consideration of the license, consent, and confidential status of the data. While experimenting with openly licensed datasets may not prohibit such use, deploying an off-site LLM to process closed (or not-yet-open) peer-reviewing data raises ethical and legal concerns.

\paragraph{Example: Data collection at ACL Rolling Review.} One example of a peer-reviewing data collection process that takes confidentiality, personal data, and licensing into account is the ``Yes-Yes-Yes'' workflow used at ACL Conferences through the ACL Rolling Review system. This opt-in workflow was developed following an in-depth community discussion and legal advice, and consists of three steps (1-3). Before reviewing starts, the reviewers are (1) asked to give informed consent for all their peer reviews in a given reviewing campaign, and to transfer a Creative Commons Non-Commercial license\footnote{\url{https://creativecommons.org/licenses/by-nc/4.0}} for their review reports to an external copyright holder (the Association for Computational Linguistics). The reviewers are given an option to request attribution: in this case, their name appears on the copyright notice, by which the reviewers get credit for their work, but effectively become deanonymized. This constitutes the ``Yes'' from the reviewers, and if reviewers do not give the permission for data use, the data collection for the given review stops. After this, the reviewing campaign proceeds, the decisions are taken, and some papers are (2) accepted. This constitutes the ``Yes'' from the PCs: an accepted paper with its intellectual content will appear in proceedings. For those accepted papers, the authors are then asked to (3) consent to the data associated with their manuscript being collected and made public. This is the ``Yes'' from the authors. In the end, only the data for accepted papers where \textit{both} reviewers \textit{and} authors have \textit{explicitly} consented to its public research use is made open. This highly selective procedure is automated and integrated into the OpenReview CMS used by ACL Rolling Review. It is supplemented by license agreements and consent forms. In the reviewing campaigns of September to November 2021, over 1,900 reviewers and the authors of 235 submissions contributed their data, later published as part of the NLPeer corpus \cite{dycke-etal-2023-nlpeer}. The workflow is deployed in the peer-reviewing system of ACL Rolling Review, collecting new data on a bi-monthly basis, which is later released to the public. The procedure itself is further developed to include additional data types and to increase author and reviewer participation. The above workflow can serve as a starting point for future data collection efforts in other communities; for further details see \cite{dycke2022yes}.

\section{Measurement and Experimentation}
\label{sec:resource_eval_eval}

How do we know whether our tools to support peer review work well? Empirical study of NLP assistance for peer review requires systematic experimentation, which amounts to measuring the \textit{intrinsic performance} of NLP systems (e.g., \textit{``does my tool reliably detect unsubstantiated claims in the manuscript?''}), as well as measuring the \textit{downstream effects} of the NLP assistance on the process (e.g., \textit{``does this tool, once deployed, reduce the time and increase the quality of peer reviews?''}). Yet, many operations and assessments in peer review are hard to formalize, which translates into the challenge of converting these operations into robust NLP tasks and evaluation criteria. This poses a major problem for NLP and AI, whose methodology largely depends on objective and measurable performance. Even with measurements well-defined, peer review remains a complex process which involves many factors and has many interacting parts. This introduces the challenge of designing experimental setups that allow measuring the effects of interest while avoiding confounds. 

\subsection{Defining and Measuring the Variables}

In our discussion of NLP-based peer-reviewing assistance so far, we have often assumed that the phenomenon to be modeled---or the outcome to be improved---is well-defined and easily measurable. Yet, this is often not the case, especially for complex, subjective phenomena like ``peer-review quality'', ``manuscript quality'' or ``reviewer expertise''. While one might be tempted to focus on easily accessible signals like ``review helpfulness score'' for review quality or ``confidence  score'' for reviewer expertise, it is important to remember that these signals only serve as imperfect proxies of the complex processes that we aim to study and improve~\cite{wang2020debiasing,goldberg2023peer}. Lack of clarity on what is measured and how it is measured can be problematic. It can affect both the direction of subsequent research (\textit{``manuscript quality prediction is solved''}), and the perception of the results by the public and the policymakers (\textit{``we can use AI to predict manuscript quality''}). Here we warn against hastily defined and poorly measured variables in NLP research for peer review, and outline a few key considerations that we deem important. Our discussion below is broadly applicable, in line with the trend towards more reliable operationalization and measurement in NLP in general \cite{belz-2022-metrological, belz-etal-2023-non, Schuff_Vanderlyn_Adel_Vu_2023, xiao-etal-2023-evaluating-evaluation}. The ideas expressed in these works are, in turn, inspired by the practices in other disciplines that face similar challenges, such as psychology, educational science, and sociology, highlighting the increasing need for collaboration between NLP and other disciplines.

Many variables in NLP for peer review are not hard objective properties. Instead, they are \emph{constructs}---hypothetical variables that we can't observe directly. Review quality, manuscript quality, and expertise are constructs---similar to ``language proficiency'', ``music taste'' or ``political preferences''. Since constructs cannot be observed directly, they need to be \emph{operationalized}: we must clearly define them, propose measurements to make quantitative statements about them, and assess whether these measurements are adequate. Like language proficiency tests serve as one imperfect measure for the actual ``language proficiency,'' any evaluation of ``reviewer expertise''---be it a confidence score, the ability to detect technical flaws, or something else---is similarly just a proxy. This needs to be taken into account and clearly communicated when reporting any scientific results in high-stakes application domains like peer-reviewing assistance.

To be able to clearly communicate about NLP and AI for peer review, we first require \textbf{precise and explicit definitions} of our constructs. For example, many of the approaches that we discussed in this paper aim to improve peer review quality. But do we agree on what exactly peer review quality is? Is it reviewers' ability to detect manuscript flaws and recognize the potential impact of the work? Is it the well-formedness of the resulting review report? Is it the utility of the review for the decision making? Or is it a combination of these? Clearly defining the variable can help design the measurements to estimate it, evaluate these measurements, inform downstream applications that make use of it, and foster shared understanding of the object of study among researchers.

Once defined, variables in peer review afford a wide range of \textbf{measurements}---including evaluation against a gold standard, participant surveys, time-related measurements, etc. Generalizing from the measurement-theoretical perspective on natural language generation by \citet{xiao-etal-2023-evaluating-evaluation}, the two core desiderata for any measurement are reliability and validity. A \textbf{reliable} measurement is robust to \textit{random} error. This includes measurement stability (e.g., obtaining same value on repeated measurement) and measurement consistency (e.g., across raters or data points). A \textbf{valid} measurement, in turn, is robust to \textit{systematic} error. This includes concurrent validity (the obtained measurements correlate with other measurements for the same construct) and construct validity (measurements behave in a way that reasonably reflects the construct, and does not reflect something else). We note that to judge measurement validity, we need multiple alternative ways to measure the variable of interest and other related variables. Designing accurate, valid, and reliable measurement methods is hard, and provides an important avenue for the future empirical study of peer review in the context of NLP and AI assistance, and in general.

\paragraph{Example: Review quality.} In the rest of this subsection, we illustrate the aforementioned discussion by an example of review quality. We limit our scope to the quality of the review report. The review report has two purposes: to inform the meta-reviewers about the content, merits, and issues with the work, and to ask the authors for clarifications and provide them with actionable improvement suggestions. Based on this, one can construct the quality of a review report as follows. We assume that to be helpful for the meta-reviewers, the review report should summarize the work and clearly outline its strengths and weaknesses. We assume that to be helpful for the authors, the review report should clearly outline the potential directions in which the work can be improved. We assume that a high-quality review should be clear and concise, written in a professional tone, well-substantiated, and that the reviewer should avoid heuristics and strategic behavior. Like manuscripts that typically fall into ``clear accept'', ``clear reject'', and ``borderline'' categories, peer review reports form clusters with respect to their quality. Thus, while we can expect agreement on review reports that are clearly high- and low-quality, we can also expect a broad class of borderline cases, where our indicators of review quality would be less reliable.

Following this notion of review report quality, we can discuss the measurements. We can measure structural well-formedness by manually or automatically labeling parts of a review according to their pragmatics, e.g., strengths, weaknesses or feedback to authors (Section \ref{sec:during_writing_prag})---and downscoring the review if one of the pragmatic categories is missing. We can apply manual analysis or automated tools to determine clarity, conciseness, and politeness of a review report. We can attempt to detect heuristics and strategic behavior in peer review text, along with substantiation cues like references to manuscript text and related work---and downscore reviews that are poorly substantiated and potentially biased (Section \ref{sec:during_writing_tone}). As a proxy of report utility, we can elicit numerical review quality scores from the authors and meta-reviewers, using either a single Likert scale, or a structured questionnaire like Review Quality Instrument \cite{Rooyen1999DevelopmentOT}.

We then need to empirically estimate the reliability and validity of each of these measurements. We take review quality rating as example. If we ask the authors to rate the reviews they receive, would their rating be stable, i.e., would they give the same rating when asked at a later date? Would the ratings given by the authors and meta-reviewers be consistent? Do review quality ratings relate in a reasonable way to other measurements of review quality, like well-formedness and substantiation? Do they correlate with some variables unrelated to review quality? For example, are longer reviews perceived as higher quality, and do unfavourable review scores make the authors judge reviews as lower-quality~\cite{goldberg2023peer}? Reliable and valid measurements can help us (1) design NLP assistance tools and policies that target specific issues with review report quality, (2) evaluate the effect of these tools on the quality of the review reports, and (3) evaluate the effect of reviewing policies by measuring aggregate quality of review reports in a given reviewing campaign. We note that each new measurement allows us not only to quantify a variable of interest, but also to validate the already existing measurements. Inconsistencies might indicate that our definition of peer review quality is incomplete or that our measurements need further refinement.

\subsection{Experiment Design}

As variables and measurements in peer review are diverse, so are the available ways to obtain these measurements. As a complex process, peer review affords many experimentation models, many of which go beyond the standard NLP practice. We briefly outline the core experimental setups that can be used to measure the performance and the effects of NLP assistance in peer review.

\begin{description}
\item[]{\textbf{Human-subject evaluation:}}
A natural method of evaluating the output of any NLP model is to ask people with suitable expertise to evaluate it. For instance, to assess a model that generates reviews of papers (or meta-reviews from individual reviews), researchers may be asked to evaluate the quality of these reviews~\cite{liang2023can,d2024marg,santu2024prompting}. In addition to the challenge of defining review quality, such human evaluation is prone to bias: for instance, \citet{goldberg2023peer} find that authors of the papers are biased by the positivity of reviews, while third-party evaluators are biased by the length, etc. These biases must be taken into account when interpreting the results. A further drawback of human evaluations in NLP for peer review is that they require effort and expertise, and might be challenging to reproduce \cite{belz-etal-2023-non}.

\item[]{\textbf{Gold-standard evaluation:}} An alternative approach is to carefully create ``gold standard'' ground truth. For example, to assess the capabilities of large language models in terms of evaluating papers, \citet{liu2023reviewergpt} (i) create a set of 14 papers with deliberately inserted errors, and assess if the LLMs can find these errors; (ii) create a set of 10 pairs of abstracts such that one has strictly stronger contents than the other in each pair, and assess if the LLMs can identify the stronger abstract; (iii) manually label the author checklists for a subset of papers from the NeurIPS conference, and assess if the LLM can verify them. A second example pertains to evaluating models that compute expertise of reviewers for papers. Here, \citet{stelmakh2023gold} construct a ``gold standard'' dataset by asking researchers to rank and rate papers they have read in terms of their own expertise. A third example is that of a `Chimera' test from~\citet{shah2022surveyextended}, which constructs a mashup of multiple papers to create a nonsensical paper. This paper is then used to test if proposed `automated reviewer systems' can detect its nonsensical nature. Objective evaluation and measurement of peer review variables meets two core challenges: not every variable can be objectively defined to yield a single ground truth; and even if it is defined, constructing an extensive gold dataset often requires substantial effort, thereby prohibiting large-scale experimentation.

\item[]{\textbf{Laboratory experiments:}} In some scenarios, simply asking participants to perform evaluations of the artifacts may not suffice, but instead the evaluation may warrant a closer replication of some aspects of an actual peer-review environment. If one cannot conduct the experiment in an actual peer-review process, one may replicate a part of the review process with either individual participants (e.g., as done for rebuttals by~\citealp{liu2023testing}) or groups of participants (e.g., as done for strategic behavior by~\citealp{stelmakh2021catch,jecmen2023dataset}). Such a setting is particularly well-suited for measuring time-related variables, as well as evaluating the performance of real-time reviewing assistance tools \cite{zyska-etal-2023-care, sun2024reviewflow}. A careful experimental design under this approach may also allow capturing parts of the review process that are concerned with the evaluation task and eliminate those that are not, thereby allowing to rigorously establish causal effects. 

\item[]{\textbf{Observational studies on open data:}} Publishers and platforms that make reviewing data open (\cref{sec:resource_eval_data}) make it possible to conduct large-scale observational studies~\cite{manzoor2021uncovering, 10.1162/coli_a_00455}. For example, one can learn how reviewers write peer reviews by collecting openly available data and using off-the-shelf discourse analysis tools (Section \ref{sec:during_writing}), or study the relationships between review texts and scores purely based on observation. The core challenge associated with such studies is the lack of influence over the experimental setup, which makes it more challenging to control for confounds and make causal claims. Further, even if well-suited for extracting high-level insights about reviewing processes in a given community, the data might be not representative of closed-reviewing scenarios and carry potential biases that influence the observed outcomes. Moreover, observational studies on open data are limited to the data types that are made open, and provides limited insight into the parts of the peer review not commonly recorded and archived, such as the processes of review writing, manuscript evaluation or final decision making.

\item[]{\textbf{Natural experiments:}} The policies or policy changes in peer review process can result in natural experiments. For example, several conferences in recent times have inserted some randomness in the reviewer assignment process~\cite{jecmen2020manipulation} in order to mitigate the issue of collusion rings. This randomization can now be exploited to understand various counterfactuals such as the effects of alternative assignment algorithms~\cite{saveski2023counterfactual}. A second example pertains to the ICLR conference which switched from single-blind to double-blind reviewing in 2018. This change in policy is used to investigate biases with respect to author identities in single-blind reviewing~\cite{manzoor2021uncovering}. Similar strategies can be applied if an NLP assistance tool is deployed in the reviewing system---taking into account the potential risks and harms of such application, as discussed below (\cref{sec:ethical_legal}).

\item[]{\textbf{Controlled experiments:}} The highest standard of scientific experimentation is offered by controlled experiments, where the researcher has a high degree of influence over the full experimental setting and can ensure minimal interference between the external factors and the target observations. An example of a controlled experiment is separating the participants in two groups corresponding to different reviewing conditions (e.g., with or without a certain NLP assistance tool) and measuring the effect of the condition on some variable of interest (e.g., reviewing time). Controlled experiments are increasingly used in machine learning research on peer review to understand different aspects of the process ~\cite{nips14experiment,tomkins2017,beygelzimer2021neuripsconsistency,stelmakh2023large,rastogi2024randomized}. These works can serve as a valuable source of inspiration for controlled experiments in the context of NLP assistance. A controlled experiment can yield crucial insights into the process, but one must also simultaneously ensure fairness and efficiency of the review process, for example, in consideration of the different treatment effects in randomized control trials. 
\end{description}

The choice of the experimental setup will ultimately depend on the variables that need to be measured. NLP assistance tools can both help obtain these measurements (e.g., deriving a politeness score based on the text), and be the target of the evaluation. In both cases, we stress the importance of studying the effects of domain shift on NLP assistance performance. For the results to be robust, the findings need to be validated across domains, scientific fields and communities, as well as across time (e.g., due to changes in topics and composition of the research community) and across reviewing procedures (e.g., the use of particular reviewing forms or discussion formats). While such comprehensive evaluation is hardly feasible, this highlights the need for \textbf{carefully documenting} the above aspects when performing experiments on peer review, including the overall procedure used, the community involved, as well as descriptive statistics and examples of the reviewing data. Finally, if the experiment involves human subjects, the researcher should obtain an \textbf{approval from an Institutional Review Board} (IRB) or an equivalent ethics board before conducting the experiment. Researchers are also encouraged to pre-register their experiments, via platforms like OSF (\url{https://osf.io}) and AsPredicted (\url{https://aspredicted.org}).

\section{Ethics of NLP-Assisted Peer Review}
\label{sec:ethical_legal}

Peer review is a hard, time-consuming task prone to error. The goal of NLP assistance for peer review is to make parts of the process more efficient and effective. Ideally, using NLP for reviewing assistance would reduce the number and severity of existing ethical issues in peer review, such as human biases. However, NLP and AI tools also have the potential to create \emph{new} risks and introduce \emph{new} biases. Therefore, it is essential to discuss the potential risks of using NLP for peer review, and how these risks can be mitigated through means taken by the application designers and by the community.

One central issue that accompanies the use of AI tools for decision support is the often implicit assumption that a more automated process is fairer, safer, or less biased than a manual process. Yet, the performance of AI tools can be misleading, and automated systems that seem to work well in the majority of cases can induce complacency, bringing new risks in terms of bias propagation, systematic error, outcome disparity, and safety.\footnote{For example, to aid criminologists in predicting whether a criminal is likely to commit more crimes, AI-based tools have been developed, which were subsequently themselves shown to exhibit systematic biases \cite{ProPublica}, resulting in a discussion about various definitions of bias.} As of today, we possess neither the technology capable of performing peer review in a satisfactory manner, nor the methodology to evaluate its effects on the affected community and the scientific process. 
While NLP tools can support peer review on the level of individual tasks, it is necessary to define the types of biases that these tools can exhibit, and to systematically monitor and quantitatively evaluate NLP assistance tools with respect to these biases---in addition to the tools' performance.\footnote{See \citet{chouldechova2016fair} and \citet{kleinberg2016inherent} for a discussion on trade-offs between biases in automatic systems and fairness criteria.}

Novel tools and policies can bring additional risks and raise new issues, but some of these risks are not new and already exist in the system. For example, algorithms may introduce biases, but even without algorithms, people already have biases encoded in their evaluation and opinions. Similarly, in terms of transparency, new algorithms require transparency, but the human process of decision making also requires transparency, which can be improved in existing processes as well. Therefore, the development of NLP assistance and the subsequent policy decisions should be made in consideration of the trade-off between risks, advantages, and benefits in the existing systems, and the risks, advantages, and benefits introduced by the new systems. We outline several areas of concern in NLP for peer review from an ethics standpoint.

\begin{description}
    \item[Bias: ] Deploying NLP assistance for peer review bears the risk of reinforcing and augmenting existing biases in the process \cite{shah-etal-2020-predictive}. Unfair decisions will not magically go away, and authors will always disagree with rejections~\cite[Section 4.2]{goldberg2023peer}. Groups of people or areas of research might feel excluded and this might lead to allocational harm (i.e., high-stakes impact of unfair decisions).
    NLP assistance models can pick up on idiolects of people who frequently published in the past.
    Thus, in the process of designing new systems, special attention needs to be paid to measuring and mitigating potential biases, old and new.
    
    \item[Transparency:] Given the importance of peer review outcomes for scientific process and research careers, a high level of transparency is desired. 
    Transparency of machine-assisted peer review can be increased by explicitly stating what types of NLP assistance are deployed in a given reviewing campaign, publishing detailed descriptions of the models used (e.g., on blogs) and model cards. Useful information includes: how the tools are built, deployed, what the tools are capable of---what they can and cannot do, how are the decisions made whether to use these tools and to what extent, and how is data privacy addressed. Using open-source models promotes scientific openness and reproducibility---yet increases the risks of malicious behavior by adversarial parties who can exploit vulnerabilities in these tools. Related to transparency is explainability: while explainable NLP is a major research direction, there are hardly any works that explicitly target explainability in NLP for peer reviewing~\cite{kim2023assisting}---an important area for future study.

\item[Agency:] With NLP in the loop, who is responsible for the outcome? If an NLP assistance module produces a faulty prediction, is it the fault of the developers who created it, the conference organizers who deployed it, or the users who failed to check the predictions? Given the ever-growing capabilities of NLP systems, reviewers, authors, and editors alike may be inclined to outsource their responsibility on the review to the ``machine'', e.g., by submitting texts and assessments automatically generated by LLMs with little further involvement. Such cases need to be explicitly addressed by the community policy---for example, it can be made clear that the user is responsible for the ultimate content and quality of the work. Parallel to that, it must be ensured that the mechanisms for reporting and recourse can adequately account for the consequences of NLP-assisted submission, matching, reviewing, discussion, etc.
    
\item[Privacy:] The sharing of peer-reviewing data and the use of models to assist in the peer review process could potentially be abused to infer personal information about the reviewers, threatening the integrity of the process. Following the example by~\citet{jecmen2020manipulation}, suppose the models used to assign reviewers to papers, the data pertaining to publication history of each reviewer, and all submissions are public. If the models and optimization procedure are deterministic, anyone can re-run these models to reconstruct the assignment of reviewers to papers, thereby compromising the anonymity of the review process. \citet{ding2020privacy,ding2022calibration} provide further examples. Additional concerns include reidentification (also called deanonymization) by analyzing the writing patterns from publicly available reviews and matching them to the publicly available papers. In addition to the careful legal treatment of reviewing data and its derivatives (Section~\ref{sec:resource_eval_data}), privacy-preserving techniques such as differential privacy \cite{journals/fttcs/DworkR14, klymenko-etal-2022-differential} and anonymization \cite{conf_icde_LiLV07} should be explored to minimize risks to the participants of the peer-reviewing process when sharing data. A disclaimer prohibiting the use of data and models for author profiling and reidentification should be included with the release of the respective resources. 
\end{description}

We stress that it is not possible to foresee all potential consequences of NLP assistance for peer review. Given the high-stakes nature of the process, once NLP systems are deployed, the users will likely adjust their behavior over time to gain further benefits from the system---intentionally or not. We note that this is not unique to NLP automation; deployment of \textit{any} new policy can have a similar effect. 
This highlights the importance of closely monitoring the changes in the reviewing process in response to new policies and tools.

\section{Conclusion: A Call for Action}
\label{sec:cfa}
Peer review is a critical part of modern science. As science accelerates, peer review faces new challenges related to logistics, cost, bias, low-quality reviewing and strategic behavior. In this white paper, we have discussed the potential ways in which natural language processing can help make the peer-reviewing process more efficient and robust, while taking into account potential legal, methodological, and ethical challenges. We believe that this research area has great potential to contribute to the study and advancement of the scientific process in the age of AI. The natural language processing and machine learning communities are uniquely positioned to study AI for peer review, as we practice peer review ourselves (and thus can reap the benefits of AI assistance), establish our own reviewing policies, and have the capability to develop tools that can alleviate some of the pressing issues with the process. We thus finish the paper with a call for action---the steps that individual readers of this work can take to help advance NLP for peer review.

In their role as \textbf{scientists}, we invite the readers to \emph{participate in the discussion}. The goal of machine-assisted peer review is to make the lives of researchers better---and this can not be done without taking the voice of the community into account. Openly available reviewing data is scarce. If you can, consider \textit{donating your data for research use}---naturally, given that the data is collected according to adequate standards of licensing, personal data and privacy protection. NLP assistance for peer review seeks to help experts. Experts are hard to recruit on crowdworking platforms and to capture in public polls. When possible, \emph{participate in the controlled experiments}, \emph{respond to surveys}, and give feedback on the deployed assistance tools.

We invite \textbf{NLP and AI researchers} to \emph{contribute work}. Peer review offers a plethora of applications for any area of NLP---from summarization to ethics to applied user studies. To help researchers get started, our companion repository (\url{https://github.com/OAfzal/nlp-for-peer-review}) compiles a set of relevant datasets pertaining to peer review. It will be regularly updated and welcomes new contributions. We call NLP and AI researchers working in the peer-reviewing domain to \emph{observe good scientific practice}. This includes responsible handling of peer-reviewing data, methodological clarity and rigor, attention to the ethical implications of the work, and honesty in communicating the results and outlining the risks. Finally, \emph{help build the community}. NLP for peer review is an emerging topic that attracts interest both within and outside NLP. Promote relevant existing work, participate in related shared tasks, join non-AI meetings on related topics,\footnote{Such as the International Congress on Peer Review and Scientific Publication, ~\url{https://peerreviewcongress.org}.}
 and consider organizing new meetings that will help grow and consolidate the community.
 
We call on \textbf{policymakers for peer-reviewed venues} to consider the opportunities of NLP assistance in peer review, but also to account for the risks. Fostering a culture of responsible tool use requires new policies. We believe four aspects to be of particular importance. First, machine-assisted quality control in science calls for \emph{transparency}. It must be clear who decides to use what tool and for which purpose, how these tools are developed, and what risks they bear. Second, it is important for the community to have trust in the tools they are using. Thus, \emph{explainable} methods for reviewing support should be given preference. Next, deploying NLP assistance in live reviewing systems calls for a high degree of quality control: new features should be rigorously tested, and processes must be in place to revert the use of new tools. This calls for careful, \emph{incremental deployment} of NLP assistance for peer review: selecting new features carefully, observing their impact and the way they affect the community, and evaluating them before introducing the next bit of technology. Finally, deployments of NLP assistance (as well as any other systematic change to the process) call for careful \emph{monitoring}, and all participants of peer review---authors, reviewers, meta-reviewers and program chairs---should be given clear \emph{feedback channels} that will help inform further policy changes. 

Finally, we invite the readers affiliated with \textbf{research-sponsoring organizations} or serving as their reviewers to prioritize research questions and investments that will \emph{enable progress in NLP and AI research for scholarly peer review}. 
As science grows, peer review in many disciplines and research communities faces challenges. Inefficient peer review diverts the public resources from producing new research and training the next generation of scientists. Unreliable peer review, in turn, threatens the integrity of the scientific process and undermines public trust in scientific work. We believe that the NLP and AI research communities should be at the forefront of developing and using tools to support peer review, and should be among the beneficiaries of the recent progress in AI that they have enabled.

\section*{Contributions}
\begin{description}
    \item[] \textbf{Coordination}: Ilia Kuznetsov;
    \item[] \textbf{Main text}: 
Osama Mohammed Afzal, 
Koen Dercksen, 
Nils Dycke, 
Alexander Goldberg, 
Tom Hope, 
Dirk Hovy, 
Jonathan K. Kummerfeld, 
Ilia Kuznetsov, 
Anne Lauscher, 
Kevin Leyton-Brown, 
Sheng Lu, 
Mausam, 
Margot Mieskes, 
Aurélie Névéol, 
Danish Pruthi, 
Lizhen Qu, 
Anna Rogers, 
Roy Schwartz, 
Nihar Shah, 
Noah A. Smith, 
Thamar Solorio, 
Jingyan Wang, 
Xiaodan Zhu.
    \item[] \textbf{Repository}: 
Osama Mohammed Afzal, 
Nils Dycke, 
Ilia Kuznetsov, 
Sheng Lu, 
Aurélie Névéol, 
Nihar Shah.
    \item[] \textbf{Review and feedback}:
Nils Dycke, 
Iryna Gurevych, 
Ilia Kuznetsov, 
Anne Lauscher, 
Kevin Leyton-Brown, 
Margot Mieskes, 
Anna Rogers, 
Nihar Shah, 
Noah A. Smith, 
Jingyan Wang.
    \item[] \textbf{Organisation}:
Iryna Gurevych,
Anna Rogers, 
Nihar Shah.
\end{description}

\section*{Acknowledgments}
The work of Nihar B. Shah and Alexander Goldberg was supported by NSF 1942124 and ONR N000142212181. The work of Ilia Kuznetsov, Nils Dycke, Sheng Lu and Iryna Gurevych was supported by the LOEWE Distinguished Chair “Ubiquitous Knowledge Processing” (LOEWE initiative, Hesse, Germany) and co-funded by the European Union (ERC, InterText, 101054961), the German Research Foundation (DFG) as part of the PEER project (grant GU 798/28-1) and by the German Federal Ministry of Education and Research and the Hessian Ministry of Higher Education, Research, Science and the Arts within their joint support of the National Research Center for Applied Cybersecurity ATHENE. The work of Kevin Leyton-Brown was funded by an NSERC Discovery Grant and a CIFAR Canada AI Research Chair (Alberta Machine Intelligence Institute). The work of Anne Lauscher is funded under the Excellence Strategy of the German Federal Government and the States. Mausam was funded by IBM-IITD AI Horizons network, grants from Google, and a Jai Gupta chair fellowship. 
The authors would like to thank Bahar Mehmani and Dennis Zyska for their valuable input and feedback. Finally, the authors would like to thank David Kunz, because ``\textit{why wouldn't they? Nobody is complaining.}''

\bibliographystyle{apalike}
\bibliography{biblio}

\end{document}